\documentclass[11pt]{article}
\usepackage[top=2cm, bottom=2cm, left=2cm, right=2cm, heightrounded,
  marginparwidth=2.9cm, marginparsep=2mm]{geometry}

\usepackage{lineno,hyperref}
\modulolinenumbers[5]
\usepackage{multirow}
\usepackage{amsmath,graphicx}
\usepackage{amssymb}
\usepackage{hyperref}
\usepackage{comment}
\usepackage{tabularx,booktabs}

\usepackage[usenames, dvipsnames]{color}

\DeclareMathOperator{\E}{\mathbb{E}}
\DeclareMathOperator*{\argmax}{arg\,max}

\usepackage{xcolor}
\newcommand\gk[1]{{\color{black}{#1}}} 
\newcommand\hw[1]{{\color{black}{#1}}} 

\title{Anomaly Detection of Adversarial Examples using Class-conditional Generative Adversarial Networks}

\author{Hang Wang, David J. Miller and George Kesidis\\
School of EECS, Pennsylvania State University, University Park, PA, 16802, USA\\
\{hzw81,djm25,gik2\}@psu.edu
}

\begin{document}

\maketitle

\begin{abstract}
Deep Neural Networks (DNNs) have been shown vulnerable to 
Test-Time Evasion attacks (TTEs, or  adversarial examples), which, by making small changes to the input, alter the DNN's decision. We propose an unsupervised attack detector on DNN classifiers based on class-conditional Generative Adversarial Networks (GANs). We model the distribution of clean data conditioned on the predicted class label by an Auxiliary Classifier GAN (AC-GAN). Given a test sample and its predicted class, three detection statistics are calculated based on the AC-GAN Generator and Discriminator. Experiments on image classification datasets under various TTE attacks show that our method outperforms previous detection methods. We also investigate the effectiveness of anomaly detection using different DNN layers (input features or internal-layer features) and demonstrate, as one might expect, that anomalies are harder to detect using features closer to the DNN's output layer.
\end{abstract}

\section{Introduction}
\label{sec:intro}

In recent years, vulnerabilities of Deep Neural Networks (DNNs) have attracted significant attention. 
It has been shown that, for a given test image to be classified, 
a malicious attacker can alter the image with small, often
imperceptible perturbations that induce the DNN to change its decision \cite{goodfellow2014explaining}. This is called a Test-Time Evasion attack (TTE) or an ``adversarial example" (AE). It is important, both in terms of security and to increase trust in and reliability of DNN decision-making, to understand TTEs and to devise defenses against them. 

Several works \cite{nguyen2015deep,hendrycks2016baseline, xiang2022post, xiang2020detection} have shown that DNNs may overfit to the training set, and thus may not generalize well on out-of-training-distribution test samples. Accordingly, various schemes have been proposed to either detect such samples and/or to robustify the DNN. We categorize these methods as follows: i) adversarial training; ii) ensembles of models; and iii) anomaly detection. Adversarial training methods \cite{madry2017towards,li2022semi} create adversarial examples (labeled to their true class of origin) and include such supervising examples in classifier training. Adversarial training 
improves robustness 
against specific attacks;
however this is achieved with some reduction in accuracy on clean data,  i.e., with an increase in the classifier's {\it bias} \cite{tsipras2018robustness}. 
Moreover, adversarially trained DNNs may still be vulnerable to {\em other} (unknown) TTE attacks that were not included in the training set \cite{carlini2017adversarial,athalye2018obfuscated}.  Thus,
such training may not resolve the fundamental problem.
Other works propose a Neural Network ensemble defense  \cite{tramer2017ensemble,hansen1990neural}, wherein several sub-networks are trained, with final decisions made by voting/aggregation of sub-network decisions. However, adversarial examples have been shown to transfer well between models trained on the same dataset \cite{tramer2017ensemble,demontis2019adversarial}. There are recent attempts trying to diversify the sub-models, to make their decisions less redundant. But those strategies may fail when the perturbation size of the attack is increased
\cite{ilyas2019adversarial}.

In this paper, we focus on an anomaly detection strategy -- instead of trying to robustly classify any given input, we aim to detect if a given input is anomalous \cite{tramer2021detecting}. If an input sample is detected, it can either be flagged with a 
``don't know" decision or alternative, robust decisionmaking may be invoked for such samples, based on the class-conditional detection statistics \cite{lee2018simple,miller2019not}.

In  \cite{nesti2021detecting,cohen2020detecting,abusnaina2021adversarial,tian2021detecting,yin2019gat}, adversarial examples are assumed to be known by the defender. A detection model (e.g., logistic regression model) is then learned to discriminate between normal and adversarial examples. However, there are numerous algorithms for creating adversarial examples --  the (supervised) detection model learned based on some (known) attacks may not generalize well for adversarial examples generated by other (unknown) attacks, not seen in the training data. \cite{grosse2017statistical} shows that adversarial examples follow a different distribution from clean samples, which inspires works that rely only on the clean dataset to model the clean distribution, and which thus utilize statistical hypothesis testing (statistical anomaly detection) to identify adversarial examples as out-of-distribution data. The class-conditional distribution of the DNN's penultimate layer features can be captured by, {\it e.g.}, Kernel Density Estimation (KDE) \cite{feinman2017detecting} or by a multivariate Gaussian Distribution \cite{lee2018simple}.
\cite{miller2019not} uses Gaussian Mixture Models (GMMs) as a more powerful distribution estimator and assesses multiple internal layers' features to perform detection. Their work proposes a novel Kullback-Leibler detection statistic, assessing discrepancy between the DNN’s class decision posterior and a class posterior formed based on null modeling of internal layer activations. \cite{miller2019not} 
reports good detection performance  
when adversarial perturbations are small, i.e., for low-confidence attacks.
However, their detector  
was not assessed for larger adversarial perturbations (high-confidence attacks).

A key question is which internal layer's activations to use in performing detection.
It is difficult to model the low-level features (i.e. features from layers close to the input layer, or the input layer itself) by a GMM or KDE model due to the non-linearity and high-dimensionality of the feature space (see (a1) and (a2) in Figure \ref{fig:viz}). The high-level features (i.e., features from layers close to the output layer) are more easily 
modelled,  but these
features may exhibit less atypicality, conditioned on the DNN's decided class, than features closer to the input layer (see (a4) in Figure \ref{fig:viz} -- the attacked `cat' images, which are original images from class `cat' but which are classified as `dog' after adversarial perturbation, are close to the clean `dog' images' distribution).
In our work we aim to develop a  model to  estimate the clean distribution of 
low-level features (or of the input image itself).

The development of Generative Adversarial Networks (GANs) gives an alternative (essentially non-parametric) way to model data distributions,
which may suffer 
less model bias than traditional density modeling methods such as
as GMMs and KDE.
Various GANs models have recently been used for anomaly detection \cite{schlegl2017unsupervised,akcay2018ganomaly}. Given a test image, the GAN's Discriminator output and the GAN's Generator Image Reconstruction Error have been used as detection statistics, where the Discriminator output is the probability
that the sample comes from the training distribution.  A small probability thus indicates an outlier sample.
The Image Reconstruction Error is the minimum squared error between Generator-reconstructed images and the test image -- anomalous test data tend to incur higher Generator Reconstruction Errors than images that follow the training distribution \cite{schlegl2017unsupervised}. The methods in \cite{schlegl2017unsupervised,akcay2018ganomaly} were originally proposed for detecting nodules in CT scans of the lung and macular degeneration in OCT scans of the retina. It is a non-trivial problem to successfully apply GANs to detect adversarial examples. Conventional GANs based detection methods are not very effective for this task because the perturbations of an attack image are often not detectable by the Discriminator, and do not significantly influence the reconstruction loss (as will be seen by our experimental results in the sequel). 



In the following, we propose a novel GANs approach for unsupervised adversarial example detection.  Our method does not require any knowledge of attacks and can exploit features at any layer of a DNN, including the input layer.  Our approach overcomes limitations of existing GANs methods for attack detection by learning class(decision)-conditional GANs models.
Our approach is illustrated in Figure 1, considering both the cases of a clean (attack-free) image and an attacked image.  In the former, the discriminator's outputs are high (indicating that this image is a plausible image and from the decided category), while the generator's reconstruction error is small.   
In the latter case, the discriminator's outputs are low (indicating that the image is implausible, i.e., inconsistent with the training distribution), while the generator's reconstruction error is large.

Our contributions are summarized as follows: 
\begin{enumerate}
\item We propose an advanced class-conditional
GAN model for anomaly detection of TTEs,
in particular AC-GAN.
\item We investigate using DNN features from different layers and find (unsurprisingly) that features extracted from layers closer to the DNN's output are less useful for detecting TTEs. 
\item Our model can correct the DNN's prediction for images that are detected as attacked -- it provides an alternate prediction based on the GAN's class conditional statistics. The code for our work is available at \cite{AC-GAN-ada-github}.
\end{enumerate}

\begin{figure}
\vspace{-1.5cm}
\begin{minipage}[b]{1\linewidth}
  \centering
  \centerline{\includegraphics[width=10cm]{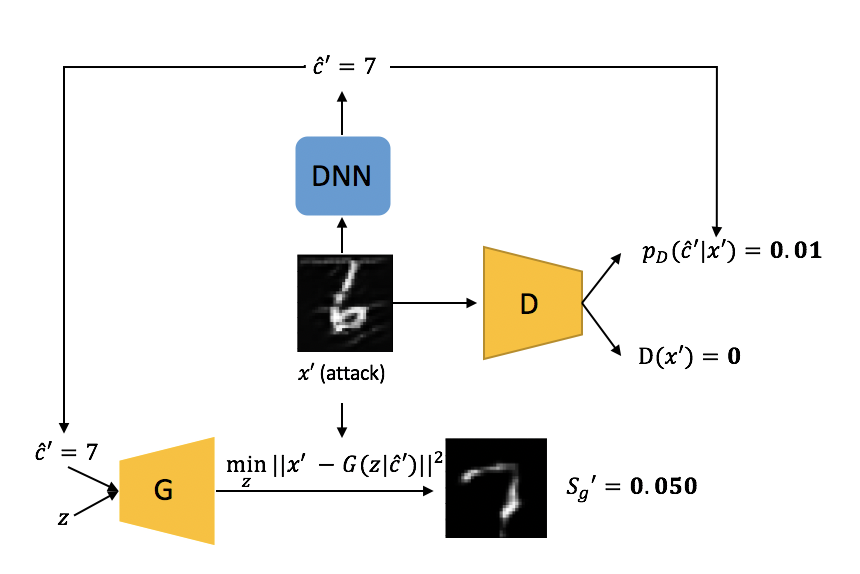}}
  
  \centerline{(a) Detection for an attack image}\medskip
\end{minipage}
\hfill
\begin{minipage}[b]{1\linewidth}
  \centering
  \centerline{\includegraphics[width=10cm]{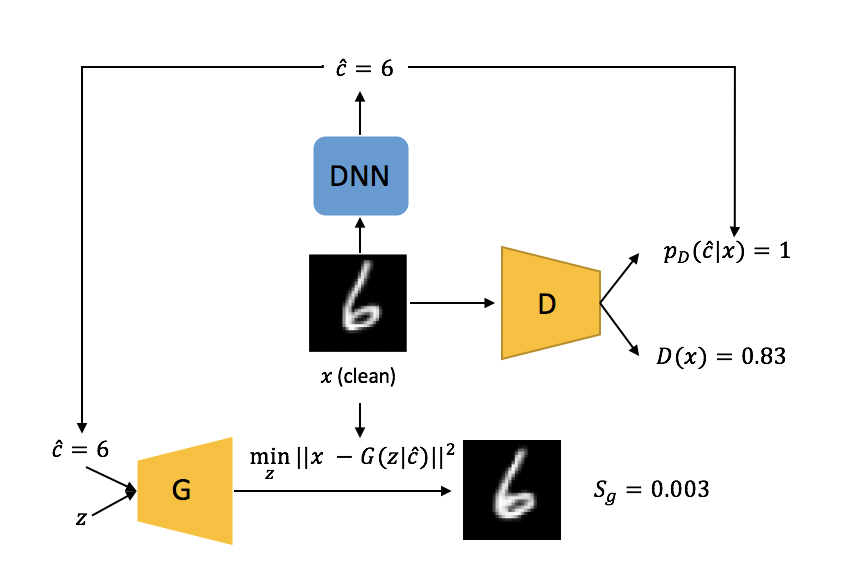}}
  
  \centerline{(b) Detection for a clean image}\medskip
\end{minipage}
\caption{Overview of our AC-GAN TTE detection method. (a)The method applied to an attacked image $x'$. The conditional distribution is modeled by an AC-GAN model. Block $G$ represents the Generator, $D$ the discriminator, and the DNN is the classifier being attacked. Given $x'$ (originally from class `6' and perturbed to class `7') and the DNN's predicted label $\hat{c}'$, the Generator is used to reconstruct image $x'$ conditioned on class $\hat{c}'$ by minimizing the square error to obtain the Generator's statistic $S_g'$. The outputs of the AC-GAN's Discriminator -- $D(x')$, the probability that $x$ follows the training distribution, and the posterior probability of the decided class, $p(\hat{c}'|x')$ -- are two other detection statistics.  Note that the Discriminator's two probabilities are low, indicating that the given image is not plausible.  Likewise, the generated image looks like a `7', not a `6' -- the associated Generator square error is large. (b) The method applied to a clean image.
 }
\label{fig:overview-attack}
\end{figure}

\section{Related Works}
\label{sec:back}

\subsection{Test-Time Evasion Attacks a.k.a. Adversarial Examples}
For image domains, a TTE
attacker aims to find a perturbation $\delta$ of a 
correctly classified
input image $x$, yielding $\hat{x} = x + \delta$, such that:
\begin{itemize}
\item  by its appearance,
$\hat{x}$ still (to a human observer) belongs to the same class as $x$, but
\item  the classifier now outputs an incorrect decision $\hat{c}(\hat{x}) \neq \hat{c}(x)$;
\item  the most potentially damaging attacks are {\it targeted}, wherein the decided class $\hat{c}$
is chosen by the attacker (and thus is not merely a misclassification). 
\end{itemize}

Numerous TTE attacks have been proposed in the literature, e.g., \cite{bai2020targeted, goodfellow2014explaining, carlini2017towards,sun2020towards, dia2021localized}.
These methods can be categorized based on the attacker's assumed knowledge.
In white-box attacks, the attacker has full knowledge of the classifier {\em and} of any defense mechanism -- such attacks could be 
consistent with an insider threat. 
In black-box attacks, the attacker does not have any knowledge of the classifier or of any mounted defense;  however,
the attacker can query the classifier to learn about it.
Here, as commonly considered in previous works \cite{miller2019not,schlegl2019f,raghuram2020detecting}, we primarily assume a threat model in between
these two (black-box and white-box) extremes. 
That is, we assume a
grey-box scenario where
the attacker has full knowledge of the classifier, but no knowledge of any defense supporting the classifier. However, we will also evaluate our method against a white-box attack.

In the following, we will evaluate defenses against two targeted TTE attacks -- the Fast Gradient Sign Method (FGSM) \cite{goodfellow2014explaining} and Carlini-Wagner (CW) \cite{carlini2017towards}. 
FGSM is a computationally efficient attack method where the perturbation $\delta$ is calculated as:
\begin{equation}\label{FGSM}
    \delta = - \varepsilon \cdot
    {\rm sign} (\nabla_x J(x, t)),
\end{equation}
where $\varepsilon$ is the perturbation size, $t$ is the target class, $J(x, y)$ is the loss function given image $x$ and class $y$, 
and {\rm sign} is an element-wise 
sign operation
($\rightarrow \pm 1$) on the gradient.
Cross-entropy loss is commonly used.  
CW learns the perturbation based,
e.g., on the following optimization problem:
\begin{equation}\label{CW}
    \operatorname*{argmin}_\delta  ||\delta||_p + c \cdot f(x + \delta),
\end{equation}
where $f(x') = max(max\{Z(x')_i: i \neq t\} - Z(x')_t, -\kappa)$
, $x$ is a correctly classified
clean sample of the attacker's source class,
$t$ is the attacker's target class,  $Z(x')_i$ is the DNN's output before softmax for class $i$. $\kappa>0$ is a hyperparameter which controls the confidence of the attack, $\|\delta\|_q$ is the $L_q$ norm (size) of the perturbation $\delta$, and $c>0$ controls $\delta$'s size.
In our experiments, the $L_2$ (Euclidean) norm was used by the CW attacker.

\subsection{Anomaly Detection Against Adversarial Examples}

\subsubsection{Supervised Detection of Adversarial Examples}

One existing group of defenses assumes knowledge of some attack methods, and treats the attacked images (generated by these known methods) as negative samples, used to train a binary classifier (attack vs. no attack).  At test time, this binary classifier is used to discriminate between clean and adversarial images.  Examples of such works include \cite{Grosse} and the supervised approach in \cite{AD}.
\cite{Li_ICCV} collected features from different internal DNN layers and then applied a multi-stage classifier, with each stage working on features from a given internal layer, to detect attack images. An attack image is identified if any of the stages decides the image is attacked. 
\cite{Metzen} also utilized supervised detectors to identify attack images based on different internal layer features, while a dynamic adversarial training process is applied to train the detector to enhance its performance.  \cite{Safety-net} developed a Support Vector Machine (SVM) model to perform detection based on  quantized internal layer features. 
\gk{\cite{ma2018characterizing}}
proposed to use a test sample's $K$ nearest neighbors to estimate the Local Intrinsic Dimensionality (LID), with a detector trained based on LID.
\cite{cohen2020detecting} proposed to use an influence function to identify, for a given test sample, the top $M$ ``most relevant'' and ``least relevant'' training samples from the DNN's decided class (with these determined based on distance to the test sample). The distances between the identified training samples and the test example (and the ranks of the distances among all the training examples) are used as the detection features and fed into an LR classifier. In \cite{abusnaina2021adversarial}, a test example's $K$ nearest neighbors, drawn from a reference set consisting of both clean and adversarial examples,  are represented as a Graph. The $K$ nearest neighbors graph is then fed into a Graph Neural Network (GNN) to do detection. 
\cite{tian2021detecting} designed a dual classifier in the Weighted Average Wavelet Transform domain. The dual classifier's output is trained to imitate the original DNN classifier's output when the input is clean, while diverging from the original DNN classifier's output when the input is adversarial. 
These supervised methods, though proven to be effective against adversarial examples from known attacks (used to train the detector) may not generalize well to unknown attacks, unseen during their training \cite{Metzen,miller2020adversarial}.

\subsubsection{Unsupervised Anomaly Detection of Adversarial Examples}

In Roth et al. \cite{roth_odds}, an input image is blurred using Gaussian noise, with the detection statistic based on the change in the DNN's logits values.  Unlike \cite{Smoothing19}, \cite{roth_odds} uses a Z-score to account for class confusion.
\cite{tian2021detecting} pointed out that adversarial examples are more sensitive to the change of the decision boundary in the highly
non-convex
region -- a slight change of the decision boundary in the highly 
non-convex
region can cause changes in class-decisions of adversarial examples while the classification of normal examples will not be affected.
\cite{hu_new} uses two statistics to characterize an adversarial example:  the change in the class posterior when Gaussian noise is added to the input; and  the number of optimization steps required to perturb a test image into other classes. The latter should be most informative for ``low confidence'' attacks, for which the number of steps should be unusually small.

Other approaches
capture the class conditional distribution of the DNN's internal layer features. 
\cite{Openmax} uses the distance between a test image's internal layer features and the class conditional mean feature vector of clean data to perform detection. 
\cite{feinman2017detecting} proposed to Kernel Density model the penultimate layer feature vector to capture the class conditional clean data distribution; they also suggested a Bayesian uncertainty measure to evaluate the proximity
of an example to clean data. 
\cite{lee2018simple} captures the clean distribution of penultimate layer features using a multivariate Gaussian Model, with detection based on the Mahalanobis distance between the given test sample and the class-conditional mean vector. 
\cite{miller2019not} applied a Gaussian Mixture Model to measure the clean distribution to guarantee good representation power (Kennel Density and multivariate Gaussian are two special cases of Gaussian Mixture Model), and performed detection based on multiple internal layers' features. A
novel Kullback-Leibler detection statistic was used in \cite{miller2019not} to evaluate the discrepancy between the DNN’s class decision posterior and a class posterior formed based on null modeling of internal layer activations. 

In this paper, alternatively we propose to use a GANs model to capture the class-decision conditional distribution of clean data -- note that an advantage of a GANs for this purpose is that, unlike conventional density modelling, no density function family (e.g., mixture of Gaussians) needs to be explicitly chosen.  In this sense, GANs essentially gives ``non-parametric" modelling of the data distribution.

\subsection{GANs based Anomaly Detection}

In a GANs model,
there is a Generator $G$ and a Discriminator $D$ \cite{gans}, where
\begin{itemize}
    \item the Generator transforms a uniform
random vector $z$ into a synthetic image $x_g = G(z)$, and
\item the Discriminator takes an image $x$ as input and its output $D(x)$ is the probability that $x$ comes from the ``real" (training set) distribution, 
$p_{\rm data}$.
\end{itemize}
The parameters of the Generator and the Discriminator are jointly
learned via a (somewhat numerically complicated)
minimax optimization process:
\begin{equation*}
      \min_G \max_D
    \E_{x \sim p_{\rm data}}\log{D(x)} 
+  \E_{z \sim p_w}\log(1 - D(G(z)))
\end{equation*}
where $p_{\rm data}$ is the empirical training set
distribution and $p_w$ is typically chosen as a
uniform distribution.
It is shown in \cite{gans} that the optimal
 discriminator output $D^*$
minimizes the Shannon-Jensen metric between $p_g$ and
$p_{\rm data}$ so that, given the optimal
generator, it cannot discriminate between
generated $\sim p_g$ and ``real" $\sim p_{\rm data}$ images,
i.e., $p_g \approx p_{\rm data}$.
If $p_{\rm data}$ is complex then, with the same goal, typically a DNN for $D$ is learned instead of attempting to use this theoretical result.

 For an out-of-distribution image 
 $x\not\sim p_{\rm data}$,
 a good Discriminator will output a very low probability that the image is real. Accordingly,
 GAN based methods have been proposed for anomaly detection, e.g., \cite{schlegl2017unsupervised,zenati2018efficient,akcay2018ganomaly,schlegl2019f}. 
 Generally, GANs based
 anomaly detection methods are based on two detection statistics: Generator statistic $S_g$ and Discriminator statistic $S_d$ \cite{schlegl2017unsupervised}:
\begin{align*}
    S_g &= \min_z \|x - G(z)\|^2,\\  
    S_d &= D(x),
\end{align*}
Thus, the smaller the Discriminator statistic the more likely the sample is an outlier. 
The Generator statistic is also called the {\em reconstruction loss}: given a test sample $x$, we search for a vector $z$ in the Generator's input space to minimize the reconstruction error $\|x - G(z)\|^2$, here using Euclidean norm.  A high Generator loss is also indicative of an anomalous sample. 

Though GANs have been successfully applied to detection problems in some domains like medical images, 
the performance of existing GANs based approaches for TTE attack detection is not so good -- the perturbations of an attack image are often not detectable by the Discriminator, and may not significantly influence the reconstruction loss.  To better exploit GANs for TTE detection, we propose a detector based on {\it class-conditional} GANs models.   We hypothesize that 
it is easier to more accurately learn  class-decision conditional GANs models than a single GANs model aiming to well-represent the composite of {\it all}
classes in the problem.  This hypothesis is supported by our experimental results, which show that class-conditional GANs outperform previous
GANs approaches and achieve excellent performance in detecting TTE attacks.

\section{TTE Detection Methodology}
\label{sec:methology}

There are multiple class-conditional GANs models, e.g., Conditional GAN (cGAN) \cite{mirza2014conditional}, Auxiliary Classifier GAN (AC-GAN) \cite{odena2017conditional}, cGAN with projection discriminator \cite{miyato2018cgans}, StyleGAN \cite{karras2020analyzing}, and ReGAN \cite{kang2021rebooting}. In this paper, we will provide a benchmark study of conditional GANs based anomaly detection using AC-GAN.

\subsection{AC-GAN}
\label{sec:ac-gan}
In AC-GAN, there is a class-conditional Generator, which takes a random vector $z$ and a class label $c$ as input and outputs the synthesized image $x_g = G(z|c)$. Also the Discriminator which, for an input $x$, gives two outputs: 
\begin{itemize}
\item  $D(x)$, 
the probability that $x$ comes from the
training distribution $p_{\rm data}$ (as
for the original GANs framework), and
\item $p_D(c|x)$, the posterior probability that $x$ is from class $c$, for every $c$. 
\end{itemize}
There are two terms in the 
AC-GAN learning objective -- the ``source'' loss $L_{\rm source}$ and the ``class'' loss $L_{\rm class}$: 

\gk{
\begin{align*}
   L_{\rm source} &= \E_{x \sim p_{\rm data}}\log D(x) 
   +  \E_{z \sim p_w, c' \sim {\sf Unif}\{1,2,\ldots,K\}}\log(1-D(G(z|c')))\\
    L_{\rm class} &= \E_{x \sim p_{\rm data}}\log p_D(c|x)  
    + \E_{z \sim p_w, c' \sim {\rm unif}\{1,2,\ldots,K\}} \log p_D(c'|G(z|c')),
\end{align*}
}
where $K$ is the number of classes in the problem and $c$ in $L_{\rm class}$ is the true 
class
for data sample $x$. 
The Discriminator is trained to maximize $L_{\rm source} + L_{\rm class}$ while the Generator is trained to minimize $L_{\rm source} - L_{\rm class}$. We refer readers to \cite{odena2017conditional} for detailed explication of the training process of an AC-GAN. It has been shown that AC-GAN may suffer from gradient explosion and mode collapse\footnote{Mode collapse means the generated images have little variation (within a class, for the case of a class-conditional GANs).} \cite{kang2021rebooting}, 
and tends to generate only images that are easily classified by the discriminator (i.e., indicating that the generator is not doing a very good job) \cite{kang2021rebooting}. Moreover, when the number of classes increases, the training process becomes more challenging, with a high chance of mode collapse in the early stage of GANs learning.  
Experiments show that the performance of anomaly detection (especially Generator based anomaly detection) is limited by the quality of the learned GANs model (cf. the G-AD row in Table \ref{table:AD} and Figure \ref{fig:reconstruct}). 
However, even with these problems, we will show that our GANs based TTE detector outperforms previous anomaly detection methods, including previous GANs based methods. 

\subsection{Detection methods} 

Using AC-GAN,  we have three statistics that can be exploited for detection;
recall Figures \ref{fig:overview-attack}.
Given an input image $x$ and the DNN-predicted class $\hat{c}$ for it, these test statistics are:
\gk{
\begin{align}
    S_R & =  D(x) 
    \label{S_R-def}\\
    S_C & = p_D(\hat{c}|x) 
    \label{S_C-def} \\
    S_g & = \min_z \|x - G(z|\hat{c})\|^2 \label{S_g-def}
\end{align}
}
$S_R$ and $S_C$ are the outputs of the Discriminator, and $S_g$ is the Generator's class-conditional reconstruction error.

A detection method can be based on the statistics from either the discriminator or the generator (or from both). Accordingly, in this section we propose 3 distinct detection methods based on an AC-GAN.
A single discriminator test statistic $S_d$ can be 
obtained by aggregating $S_R$ and $S_C$ based on an independence assumption for these two probabilities, i.e., via:
\gk{
\begin{equation}
    S_d = \log D(x) + \log p_D(\hat{c}|x).   
\end{equation}
}
We call this Discriminator based Anomaly Detection (D-AD).
The first (class-independent) 
term in $S_d$ penalizes images with large perturbation size, while the second term penalizes images with low probability for the class to which the DNN is deciding. A detection is declared when $S_d$ is smaller than a given threshold.

A detection decision can also be made solely relying on the Generator, namely Generator based Anomaly Detection (G-AD),
with $S_g$ used to detect anomalous images. Again, out-of-distribution image-label pairs will give larger values of $S_g$. 

We can also combine all the test statistics to make a detection. 
For example, define the vector $S = [S_R, S_C, S_g]$. 
We suggest to model $S$ using a multivariate Gaussian Distribution, with anomalies then detected based on a p-value assessed with respect to this distribution.
We denote this method as All-AD.

Note that all of the methods require setting a threshold to make detections. Generally a threshold can be chosen in unsupervised fashion based on the largest False Positive Rate that is tolerable. In our work, we first get the detection ROC curve by conducting a grid search of threshold values; we then evaluate all the anomaly detection methods to be compared using pAUC-0.2, the partial area under the ROC curve when the False Positive Rate is less than 0.2 (since we are only interested in the True Detection Rate when the False Positive Rate is low).

\section{Experiments}

\subsection{Datasets}

We evaluated the performance of detection methods on several well-known image classification datasets:  
MNIST \cite{lecun-mnisthandwrittendigit-2010}, CIFAR-10 \cite{krizhevsky2009learning}, and Tiny-ImageNet-200 \cite{imagenet}. 


MNIST is a hand-written digit gray-scale image dataset used in classification problems. There are 10 classes corresponding to digits 0-9, with 7000 images per class. The MNIST dataset has a standard train-test split: there are 60,000 images in the training set and 10,000 images in the test set. The size of an image is 
$28 \times 28$. The CIFAR-10 dataset consists of 60000 colour images in 10 classes, with 6000
images per class. There are 50000 training images and 10000 test images. Image sizes are $32\times 32$. Tiny-ImageNet-200 is a subset of ImageNet; there are 200 classes in the dataset, with each class consisting of 500 training images and 25 evaluation images. All images are resized to $64 \times 64$.

Although DNN models have achieved high accuracy on these datasets, (e.g., ResNet-18 has 99.5\% test accuracy on MNIST and 93.02\% test accuracy on CIFAR-10 \cite{pytorch-resnet}), it has been shown that 
a correctly classified image can be easily perturbed (with a relatively small perturbation size) 
so that it is misclassified to a target class \cite{cw}.  It is also pointed out in \cite{carlini2017adversarial} that MNIST has somewhat different
security properties than CIFAR-10, with some detection methods effective on MNIST but quite ineffective on CIFAR-10. Most previous works, e.g., \cite{feinman2017detecting, lee2018simple, miller2019not}, reported experiments on both MNIST and CIFAR-10. According to \cite{carlini2017adversarial}, due to the lack of diversity of samples in MNIST, the DNN classifier trained on MNIST achieves high certainty (class decision confidence) on the test examples, which makes the perturbation size required by a successful attack larger than that for CIFAR-10. There is no detailed study on the security properties of the larger, higher resolution Tiny-ImageNet-200 dataset.

\subsection{Model architectures and Attack methods}

\begin{table}[!ht]
\begin{center}
\scriptsize
\begin{tabular}{cccc}
\toprule
        
         & MNIST & CIFAR-10 & Tiny-ImageNet  \\ \hline  

test accuracy&   99.43 \%  &  92.84\%  &  61.35\%\\

\bottomrule
\end{tabular}
\caption{Test set accuracy of DNN classifier on different datasets.}
\label{table:acc}
\end{center}
\end{table}

The ResNet-18 \cite{he2016deep} model is used as the DNN classifier. The test set classification accuracy of DNN on different datasets are shown in Table \ref{table:acc}. 
CW and FGSM are the two adversarial attack methods considered here. FGSM is a simple, one-step attack method, while CW is an optimization based method. Usually FGSM requires a larger perturbation size than CW to generate an adversarial example.

For the targeted CW attack \eqref{CW} , we set the model parameter $\kappa=14$ (which is a common setting) and searched\footnote{grid search} for the minimum value of $c$ to make successful high-confidence and low-confidence attacks. A successful low-confidence attack perturbs an image until the target-class posterior of the attacked DNN is just larger than that of any other class,  while a successful high-confidence attack is achieved when
the target-class posterior 
is larger than 90\%.
For the FGSM attack \eqref{FGSM}, we chose $\varepsilon=0.05$ for CIFAR-10 and Tiny-ImageNet-200 and $\varepsilon=0.2$ for MNIST. FGSM is a computationally efficient albeit a weak attack, so we consider FGSM attack examples with low-confidence as successfully attacked examples (i.e., we do not consider high-confidence FGSM attacks here). Some examples of the attack images can be found in \ref{appendix:attack-images}.

 We choose the correctly classified test images as the clean image set $X$, and adversarially perturb them to randomly generated target classes (different from the class labels) using a given attack method; the successfully attacked images from $X$ are denoted by $X'$. As described above, for different attack methods, we have different criteria for attack success.
The union of $X$ (i.e., the clean images) and $X'$ (the successful attack images) are used to evaluate anomaly detection performance.

We used different AC-GAN structures for the MNIST and CIFAR-10 datasets. For MNIST, there are 4 convolutional 
layers in the discriminator and 4 transposed convolutional layers \cite{radford2015unsupervised} in the generator. For CIFAR-10, both the generator and the discriminator use architectures akin to ResNet-18.
See the Appendix and  our GitHub page \cite{AC-GAN-ada-github}
for additional information regarding the DNN architectures used.
\subsection{AC-GAN training}
The objective functions defined in \ref{sec:ac-gan} are used to train the generator and discriminator. Following the training configuration in \cite{odena2017conditional}, We use mini-batch gradient descent to train the generator and the discriminator. In each mini-batch a subset of training images is sampled as real images and a set of fake images are generated by the generator given a randomly generated vector and label; then one discriminator optimization step and one generator optimization step are performed independently. A larger batch size (256 for MNIST and CIFAR-10 and 64 for Tiny-ImageNet-200) is used  to increase the training stability. To mitigate the mode collapse problem we applied label smoothing \cite{muller2019does} to the real/fake labels. The training hyper-parameters for different datasets can be found in Table \ref{table:hyper}. 

\begin{table}[!ht]
\begin{center}
\scriptsize
\begin{tabular}{cccc}
\toprule
        
         & MNIST & CIFAR-10 & Tiny-ImageNet  \\ \hline \hline 

Optimizer&  \multicolumn{3}{c}{Adam($\alpha$ = 0.0002, $\beta_1$ =0.5, $\beta_2$=0.999) }    \\
Learning rate&   0.001   &   0.0002 &0.0001    \\
Epochs  &    60    &   90         &   100         \\ 
Batch size&    256    &   256         &   64         \\ 
Weights initialization&  \multicolumn{3}{c}{Gaussian($\mu$ = 0, $\sigma$ =0.02) }       \\
Bias initialization&\multicolumn{3}{c}{0 }       \\

\bottomrule
\end{tabular}
\caption{Hyper-parameters for AC-GAN training}
\label{table:hyper}
\end{center}
\end{table}

\subsection{Performance of Anomaly Detection}
 
For TTE detection, we now report the results of the proposed D-AD, G-AD and All-AD methods, in comparison with benchmark methods. D-AD-L1 is the D-AD method with  features extracted from the first convolutional layer used for detection.  For all other variants of our method (D-AD, G-AD, and All-AD) the GANs is modeling the input (image). We compared our methods with several recent unsupervised anomaly detection methods. Kernel Density (KD) \cite{feinman2017detecting}, Mahalanobis Distance \cite{lee2018simple}, ADA \cite{miller2019not} were chosen as class-decision conditional Anomaly Detection baselines. F-anoGAN \cite{schlegl2019f} is a GAN based anomaly detection baseline. \cite{roth_odds} is a blurring based method. \hw{SID \cite{tian2021detecting} is a supervised method, where two networks are trained on the original images and the images' Weighted Average Wavelet Transforms (WAWT) respectively; the internal layer features of the two networks are fed into a binary classifier to do detection. }
We implemented \cite{miller2019not}, \cite{feinman2017detecting} and \cite{lee2018simple}; for F-anoGAN \cite{schlegl2019f}, OODS \cite{roth_odds} and SID \cite{tian2021detecting} we used the code provided by the authors. \hw{Again note that SID is a supervised method; to make a fair comparison with it, we assume the FGSM attack is known for SID when we are doing detection of CW attacks and we assume the CW low-confidence attack is known when we are doing SID detection of FGSM  attacks.}
The performance measure chosen for all the anomaly detection methods was the partial area under the ROC curve for FPR below 0.2 (pAUC-0.2). The maximum achievable value of pAUC-0.2 is 0.2, which means the method has 100\% detection rate of adversarial examples and 0 false positives (clean examples that are detected as adversarial). A larger pAUC-0.2 score means better performance.
pAUC was assessed because the most practically important detection performance is in the low false-positive regime. The experimental results for MNIST and CIFAR-10 are shown in Table \ref{table:AD}. 

\begin{table*}[!ht]
\begin{center}
\tabcolsep=0.2cm{
\scriptsize
\begin{tabular}{ccccccc}
\toprule
         & \multicolumn{3}{c}{MINST}         & \multicolumn{3}{c}{CIFAR-10}   \\ \cmidrule(lr){2-4}\cmidrule(lr){5-7} 
         & CW-HC & CW-LC & FGSM & CW-HC & CW-LC & FGSM \\ \hline \hline

f-AnoGAN \cite{schlegl2019f} &   0.0981      &0.0995        &   0.0887   &    0.0576          &    0.0563         &    0.0566  \\

KD \cite{feinman2017detecting}    &   0.1892     & 0.1887   &     0.1880  &    0.0533          &  0.0584         &  0.1642   \\

MD \cite{lee2018simple}    &     0.1861   &  0.1832  &  0.1901    &    0.1042          &   0.1125        &   0.1783   \\

ODDS \cite{roth_odds}    &  0.0618      &  0.0412 &  0.0537 &    0.0910          &   0.0568        &   0.0436   \\
SID \cite{tian2021detecting}    &  0.1576     &  0.1628  &  0.1726    &    0.1489         &   0.1412        &   0.1388   \\
ADA \cite{miller2019not}     &     0.1715   &  0.1732  &  0.1823    &    0.1593          &   0.1601        &   0.1782   \\
G-AD     &        0.1525&   0.1517     & 0.1612     &   0.0181   &    0.0254    &   0.0203 \\
D-AD     &    0.1915   &   {\bf 0.1970}   &   0.1862    &  {\bf 0.1881}      &     {\bf 0.1899}   &    {\bf 0.1819}  \\

All-AD   &        {\bf 0.1923}      &   0.1964          &    {\bf 0.1905}   &  0.1798        &     0.1825        &   0.1618   \\
D-AD-L1     &    0.1897   &    0.1873   &   0.1824    &  0.1805      &     0.1787   &  0.1768    \\
\bottomrule
\end{tabular}
}
\caption{pAUC-0.2 results of different 
detection methods under different attacks.} \label{table:AD}
\end{center}
\end{table*}

The f-AnoGAN method, which is a GAN based Anomaly Detection method but not class-conditional, has poor performance on both datasets. The blurring based method \cite{roth_odds} is also ineffective on both data sets.
Two methods based on the clean distribution of the penultimate layer features (MD and KD) have good performance on the MNIST dataset, but for CIFAR-10, especially under the CW attack, the performance is poor.  \hw{The SID method can generalize pretty well to unknown attacks, but the performance is still not as good as our D-AD.}

ADA, which assesses multiple hidden layers and uses GMMs to capture the clean distribution for these layers, outperforms all the above-mentioned methods under the two most challenging scenarios (CIFAR-10 CW-HC and CW-LC). Note that 
ADA relies on the Bayesian Information Criterion (BIC) \cite{Schwarz} to select the number of components in a Gaussian mixture model, used to represent a given hidden layer, conditioned on a particular decided class.  In order to assist BIC in choosing good model orders, we used an auto-encoding network to reduce the feature dimensionality of a given hidden layer.  The details of this approach are provided in \ref{appendix:a}.

\begin{figure}[!ht]
\begin{minipage}[b]{.48\linewidth}
  \centering
  \centerline{\includegraphics[width=6cm]{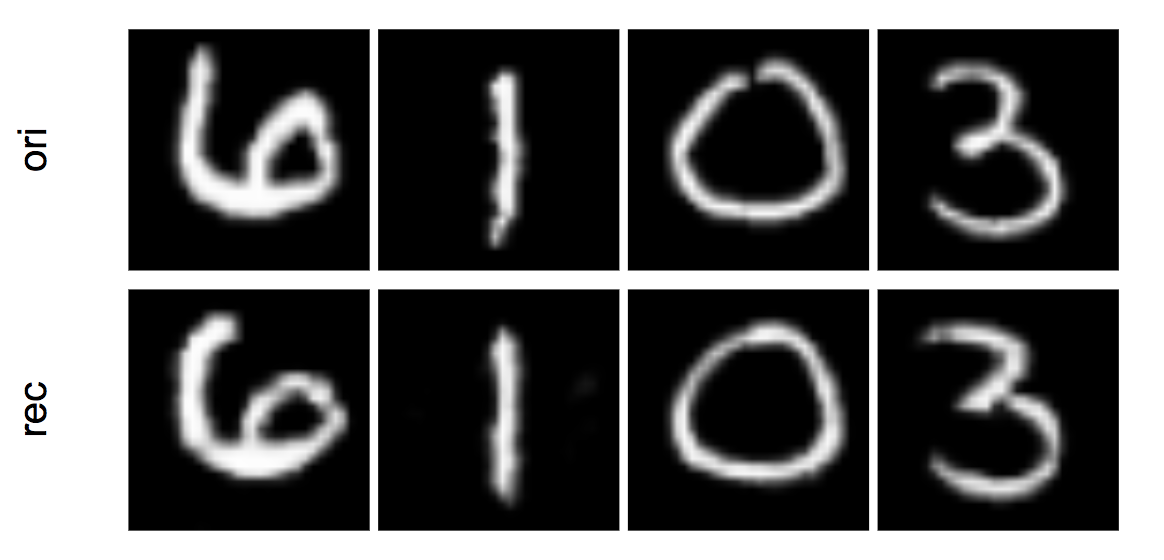}}
  \centerline{MNIST}
\end{minipage}
\begin{minipage}[b]{0.48\linewidth}
  \centering
  \centerline{\includegraphics[width=6cm]{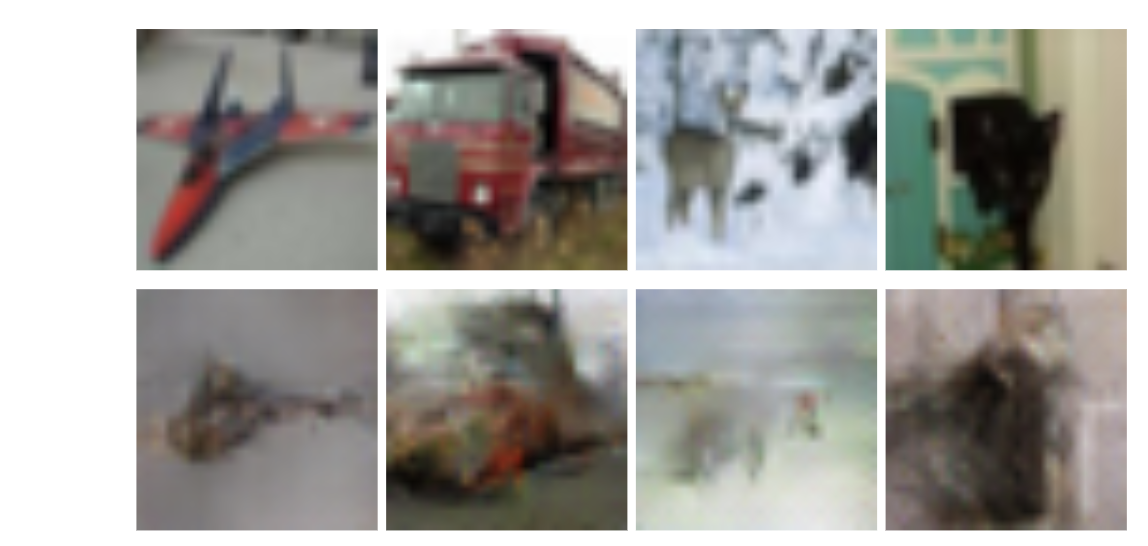}}
  \centerline{CIFAR-10}
\end{minipage}

\caption{Reconstruction examples of clean images from MNIST and CIFAR-10 datasets, the first row are the original clean test images, the second row are the images reconstructed by the Generator.}
\label{fig:reconstruct}
\end{figure}

From  Table \ref{table:AD},
we can see that our methods, which combine class-decision conditioned anomaly detection and GANs based anomaly detection, outperform both the unconditional GANs based anomaly detection method and the class-conditional ADA method. The best-performing versions of our method either use both the discriminator and generator statistics
(ALL-AD) or just use the discriminator (D-AD) -- the generator is modestly helpful for MNIST, but unhelpful for CIFAR-10. 
Notably, for the generator detection statistic $S_g$, the minimum achievable reconstruction error, we need a multiple step (usually around 300 iterations for the MNIST dataset) gradient descent process to obtain $S_g$. However for the discriminator based statistics $S_C$ and $S_R$ we only need one forward pass of the discriminator to perform detection. In our experiments we notice that the performance of the pure generator based method is very poor on CIFAR-10. Generator based detection requires accurate reconstruction of a given clean image -- for the MNIST dataset, the reconstruction task is relatively easy due to the low diversity of the data in each class.  On the other hand, the high data diversity of the CIFAR-10 dataset makes it a challenging task to reconstruct a given clean test image. \hw{Some examples of reconstructed clean images for MNIST and CIFAR-10 are given in Figure \ref{fig:reconstruct}. We can see for the MNIST dataset that the clean images can be reconstructed with high accuracy, but for the CIFAR-10 dataset, the reconstruction error is high even for clean samples.}

\begin{table*}[!ht]
\begin{center}
\tabcolsep=0.2cm{
\scriptsize
\begin{tabular}{ccccccc}
\toprule
& CW-HC & CW-LC & FGSM \\ \hline \hline 
f-AnoGAN \cite{schlegl2019f} &0.0571       &   0.0523   &   0.0655  \\
KD \cite{feinman2017detecting} &  0.0542     & 0.0567      &  0.1168   \\
MD \cite{hu_new} &  0.0918     &   0.0864   &   0.1104  \\
ADA \cite{miller2019not} &0.1312&0.1348&0.1385\\
D-AD &\bf{0.1532}&\bf{0.1585}&\bf{0.1496}\\
\bottomrule
\end{tabular}
}
\caption{pAUC-0.2 results for Tiny-ImageNet-200 dataset.} \label{table:tiny}
\end{center}
\end{table*}

\hw{The experimental results on the Tiny-ImageNet-200 dataset are presented in Table \ref{table:tiny}; we chose to evaluate our D-AD method and the other methods that have good performance on the MNIST or CIFAR-10 datasets. Our D-AD method outperforms all other 
methods on Tiny-ImageNet-200.}

\subsection{Detecting based on internal layer features}
\label{sec:internal}

Recall that previous Anomaly Detection methods, e.g., \cite{zheng2018robust,miller2019not,raghuram2020detecting, feinman2017detecting,lee2018simple}, detect anomalies based on internal layer DNN features. They model the null distribution for these features using Gaussian mixture models, kernel density estimation or $K$-nearest neighbors ($K$NN), respectively . As we discussed above, the AC-GAN model can also capture the conditional distribution of the internal layer features. 
Consider an $L$-layer DNN $h$ (e.g., of the ResNet type). For a test image $x$ with predicted class 
$\hat{c}=h_L(x)$, we first extract a feature vector from the $k^{\rm th}$ DNN layer, $h_k(x)$, which is modeled by an AC-GAN. Then the detection statistics discussed in 
\eqref{S_R-def} - \eqref{S_g-def} can be used just by replacing $x=h_0(x)$ by $h_k(x)$.
In Table \ref{table:AD}, we report the anomaly detection performance of our discriminator based GANs based detector using the features from the output of the first convolutional layer (D-AD-L1).  We can see that the detection pAUC-0.2 value is comparable to, but not quite as good as, our best-performing version. 

\begin{figure}[!ht]
\begin{minipage}[b]{.24\linewidth}
  \centering
  \centerline{\includegraphics[width=3.3cm]{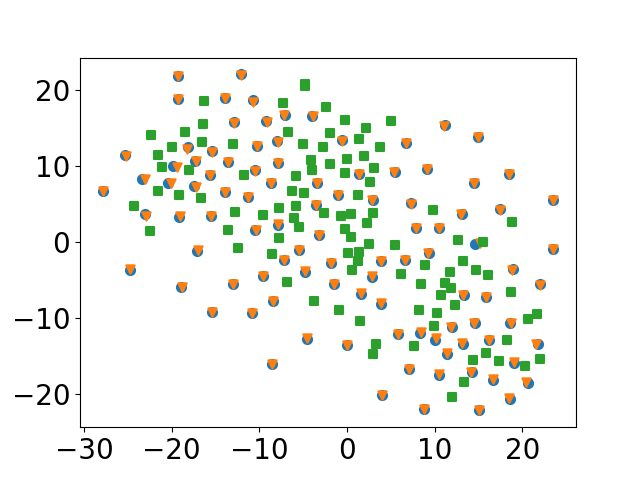}}
  \centerline{(a1)}
\end{minipage}
\begin{minipage}[b]{0.24\linewidth}
  \centering
  \centerline{\includegraphics[width=3.3cm]{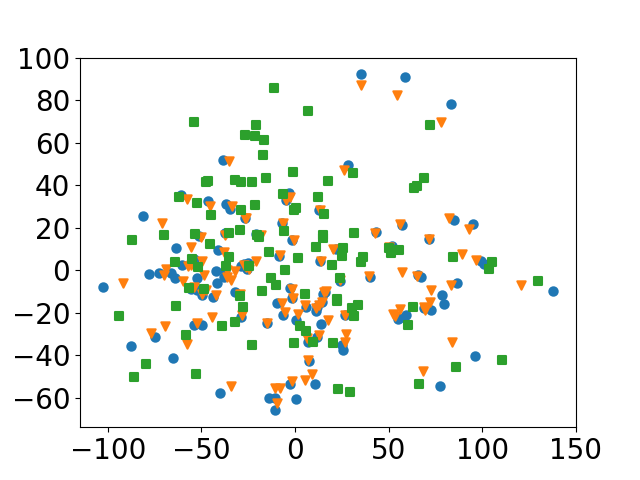}}
  \centerline{(a2)}
\end{minipage}
\begin{minipage}[b]{.24\linewidth}
  \centering
  \centerline{\includegraphics[width=3.3cm]{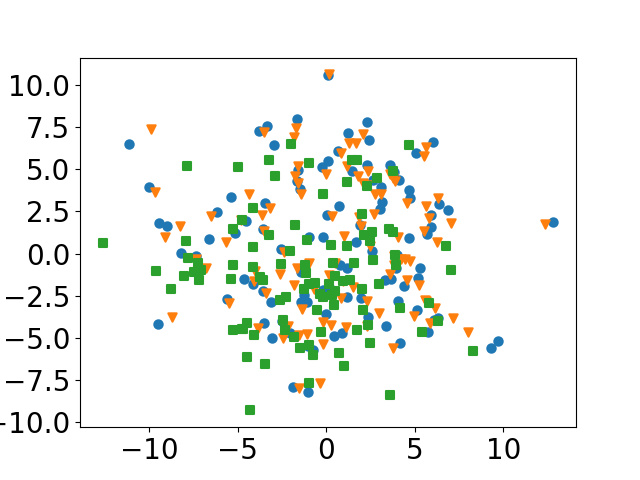}}
  \centerline{(a3)}
\end{minipage}
\hfill
\begin{minipage}[b]{.24\linewidth}
  \centering
  \centerline{\includegraphics[width=3.3cm]{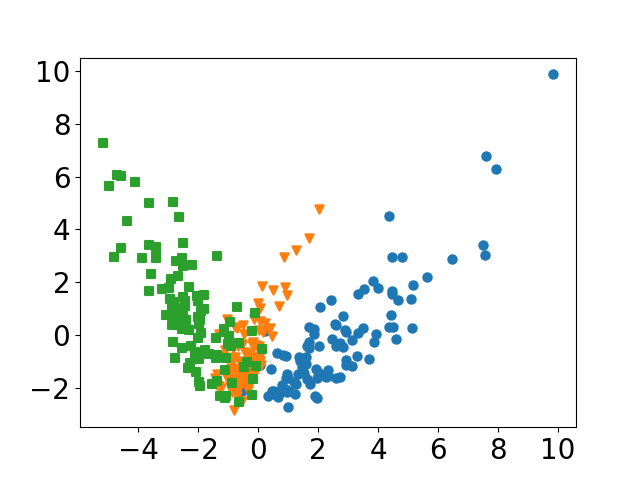}}
  \centerline{(a4)}
\end{minipage}

\begin{minipage}[b]{.24\linewidth}
  \centering
  \centerline{\includegraphics[width=3.3cm]{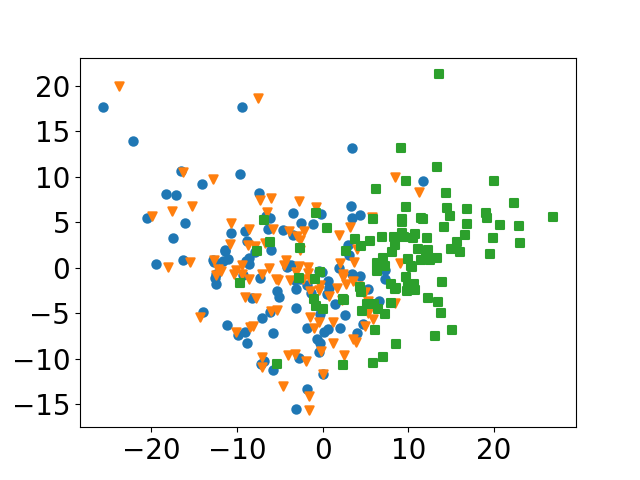}}
  \centerline{(b1)}
\end{minipage}
\begin{minipage}[b]{0.24\linewidth}
  \centering
  \centerline{\includegraphics[width=3.3cm]{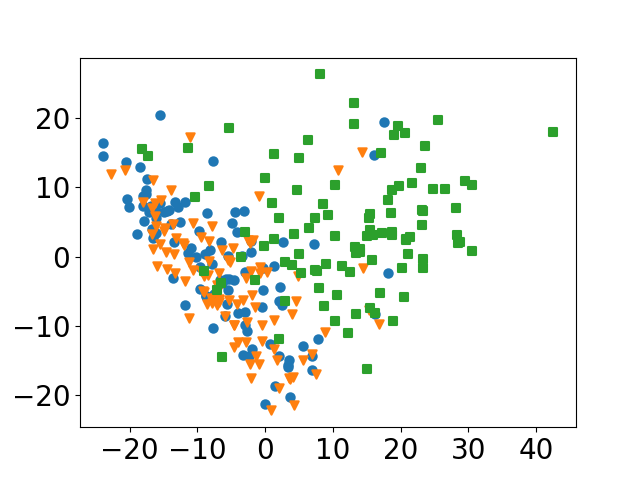}}
  \centerline{(b2)}
\end{minipage}
\begin{minipage}[b]{.24\linewidth}
  \centering
  \centerline{\includegraphics[width=3.3cm]{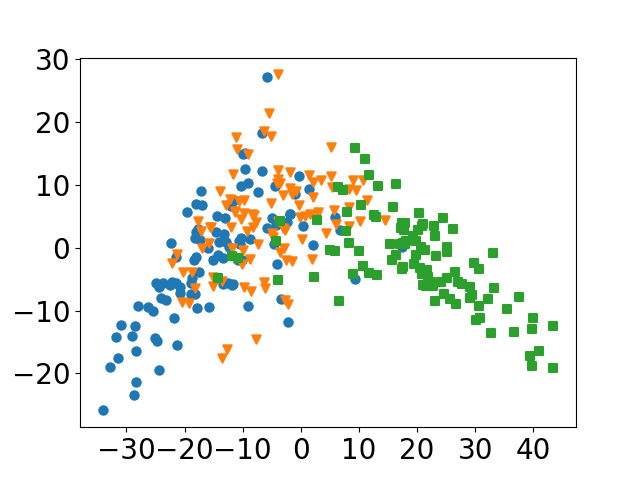}}
  \centerline{(b3)}
\end{minipage}
\begin{minipage}[b]{.24\linewidth}
  \centering
  \centerline{\includegraphics[width=3cm]{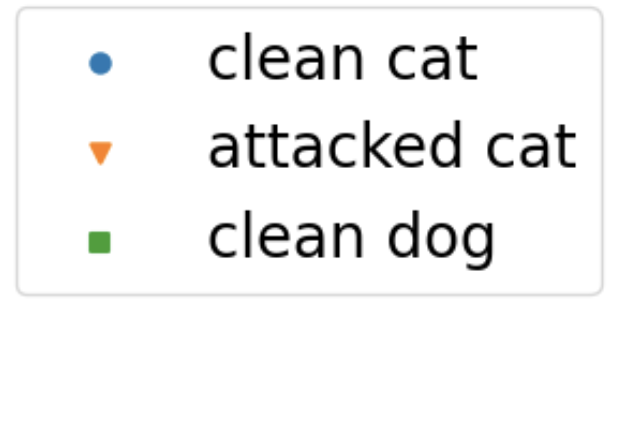}}
  \centerline{}
\end{minipage}
\hfill
\caption{Visualization of features from different DNN (ResNet) layers using the  PCA method on the CIFAR-10 dataset. The features visualized in the first row are from: (a1) the input layer  (images), (a2) the first convolutional layer, (a3)  the 
$11^{\rm th}$ convolutional layer, or (a4)  the penultimate layer.  
For sub-figures (b1)-(b3)
we train ACGAN model based on features from input layer, first  and $11^{\rm th}$ convolutional layers respectively,   features are then fed into the corresponding ACGAN's discriminator, we
extract the Discriminator’s penultimate-layer activations and visualize them. The blue circle points are clean images from class `cat', the green rectangle points are clean images from class `dog', the orange triangle points are attack images which are perturbed from class `cat' to class `dog'.}
\label{fig:viz}
\end{figure}

The AC-GAN framework also gives a way of visualizing the DNN's 
\gk{internal} features.
At test time, when an image or internal layer feature-vector is fed into the Discriminator
\gk{(depending on which layer the AC-GAN was trained on)}, 
we extract the Discriminator's penultimate-layer activations and visualize them using Principal Component Analysis (PCA). The visualization results are shown in second row of Figure \ref{fig:viz}.
Consider in particular images  perturbed from class `cat' to class `dog' (orange triangle points, attacked `cat'). In the input layer, clean `cat' and attacked `cat' are well-separated from clean `dog'.  In the first convolutional layer, attacked `cat' images still follow the clean `cat' distribution (and hence are discriminable from clean `dog'); for the $11^{\rm th}$ convolutional layer features, attacked images fall in between clean `cat' and clean `dog' \gk{(hence, just based on this low-dimensional representation, it is possible that they could be assigned to either class)}.
Finally, the penultimate layer  DNN \gk{(not Disriminator)} features (a4)  for some attacked 'dog' images, are very close to some clean 'dog' samples, which is indicative that these features, even with a powerful distribution model, cannot reliably discriminate between clean and attacked images.

\subsection{Correcting the Classifier's Prediction}
\label{robust-class}
Beyond detecting attacks,
the class-decision conditional detection statistics of the Generator and the Discriminator can also be used to {\em correct} the DNN classifier's predictions for samples that are detected as TTE attacks.  The Discriminator captures the class conditional distribution of the input data, so the class under which the input image is {\em least  atypical} can be predicted as the correct class. At test time, when an input image $x$ is detected as an adversarial example,
we can use the Discriminator's maximum {\it a posterior} class ($c_D = \argmax_i p_D(i|x)$) as the corrected class decision. Note that the DNN has a class decision $\hat{c}$ on the test image $x$, since the image is detected as an adversarial example, the class $\hat{c}$ should be an incorrect class decision, it is reasonable to exclude $\hat{c}$ when correcting labels. In practice we exclude the $\hat{c}$ when performing SoftMax.
We use only the attack images that are successfully identified by our D-AD anomaly detection method.
The corrected classification accuracy is presented in Table \ref{table:robust},
where some reported results include $\hat{c}$ and others exclude $\hat{c}$.

 \begin{table}[!ht]
\begin{center}
\scriptsize
\begin{tabular}{cccccc}
\toprule
         & & \multicolumn{2}{c}{MINST}         & \multicolumn{2}{c}{CIFAR-10}      \\ \cmidrule(lr){3-4}\cmidrule(lr){5-6} 
        &Include $\hat{c}$ & CW-HC & CW-LC & CW-HC & CW-LC \\ 
         \hline \hline 

\multirow{2}{6em}{D-GAN\cite{samangouei2018defense}} &\checkmark &   0.9247   &    0.9273         &0.2089        &0.2079             \\
  &  &   0.9314   &    0.9291         &0.3274        &0.3782             \\
\multirow{2}{6em}{D-AD}   &\checkmark &       0.8951  &   0.8816          &     0.8266    &   0.8191         \\
  &  &       0.8964  &   0.8819          &     0.8423    &   0.8316        \\
\multirow{2}{6em}{D-AD-L1}   &\checkmark &     0.8295       &    0.8204   &   0.7643        &      0.7420                \\
  & &     0.8421       &    0.8395   &   0.7874        &      0.7535                \\
\multirow{2}{6em}{pix2pix}  & \checkmark  &   {\bf 0.9504}       &    {\bf 0.9668}   &   {\bf 0.8539}        &      {\bf 0.8566}                \\
  &   &   {\bf 0.9612}       &    {\bf 0.9671}   &   {\bf 0.8632}        &      {\bf 0.8788}                \\\bottomrule
\end{tabular}
\caption{Classification accuracy in correcting the DNN decision, for correctly detected adversarial examples. CW-HC means CW high confidence attack; CW-LC means CW low confidence attack. D-GAN means Defense-GAN method.}
\label{table:robust}
\end{center}
\end{table}
 
Similar to the Anomaly Detection task, the class label correction
can be performed using a GANs based on 
features from the input layer (D-AD) or the output of the first convolutional layer (D-AD-L1).
We compare these methods with the Defense-GAN (D-GAN) \cite{samangouei2018defense} baseline. Defense-GAN is a generator based robust classification method; it uses the generator to reconstruct a given test image. The reconstructed image, instead of the original image, is then fed into the DNN classifier. 

We can see from Table \ref{table:robust} that the performance of D-GAN is better than our methods on MNIST, but our methods greatly outperform D-GAN on CIFAR-10. As discussed earlier, the performance of the generator based method is limited. For CIFAR-10, if a better generative model is used, then the generator based method may achieve better performance. We validate this hypothesis by introducing an alternative pix2pix image generation model, and present its performance for class correction in Table \ref{table:robust}. 
This method achieves the best class correction performance, amongst all the evaluated methods.
The details of this pix2pix based method can be found in \ref{appendix:b}.

\subsection{White-box attack}
We now evaluate our AC-GAN based detector under the white-box scenario, where the attacker knows everything including the DNN model, the AC-GAN defense model, the detection statistics and detection rules. In this section, we consider our best-performing version (from Table \ref{table:AD}): D-AD.  A white-box attack strategy that is widely adopted in previous works \cite{carlini2017adversarial,miller2019not, lee2018simple, ma2018characterizing} is to minimize a loss function that is a composite of the attack objective \eqref{CW} (classification loss)  and a detector loss.  That is,
\begin{equation}\label{WHITE}
    \operatorname*{argmin}_\delta  f(x + \delta) + \alpha  ( D(x+\delta) +  p_D(t|x+\delta)),
\end{equation}
where the first term is the objective of the standard CW attack  \eqref{CW}, \hw{with the second term the sum of the Discriminator's statistics. Minimizing the first term will defeat the classifier while minimizing the second term defeats the detector. 
$\alpha$ is a positive constant which was chosen by a grid search as the value that makes the most number of ``attack images'' $x+\delta$ which defeat both the classifier and the detector under the same perturbation size (a small perturbation size that guarantees not all the attack images defeat both classifier and the detector); $\alpha=5$ in our experiment.}

\begin{figure}[!ht]
\begin{minipage}[b]{.48\linewidth}
  \centering
  \centerline{\includegraphics[width=6cm]{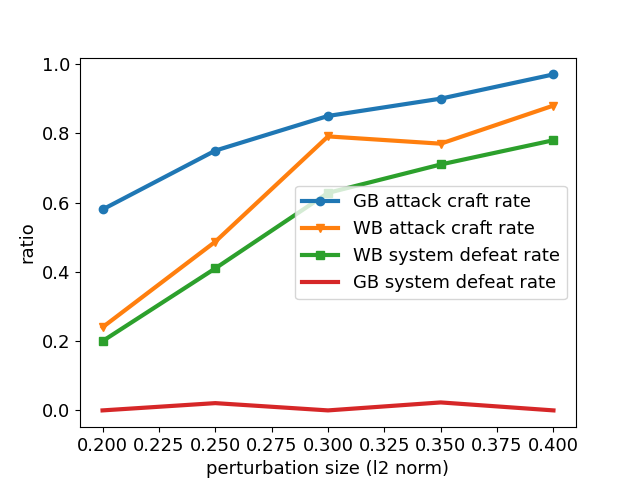}}
  \centerline{(a)  }
\end{minipage}
\begin{minipage}[b]{0.48\linewidth}
  \centering
  \centerline{\includegraphics[width=6cm]{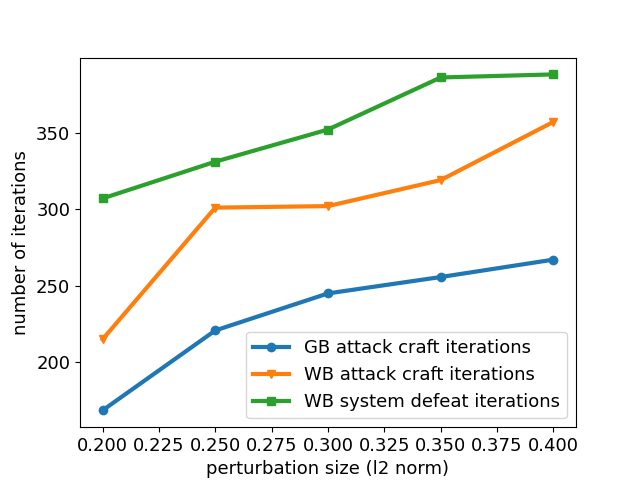}}
  \centerline{(b) }
\end{minipage}

\caption{(a): Several attack success measurements under white-box (WB) and grey-box (GB) attack scenarios. (b): Average iterations required for successful attacks. The experiments are conducted on the CIFAR-10 dataset and Projected Gradient Descent (PGD) is used to control the perturbation size.}
\label{fig:WHITE_BOX}
\end{figure}

To evaluate the robustness of our detection method, we randomly sampled 1000 test images from the CIFAR-10 test set that are correctly classified by the 
DNN classifier. We then created adversarial examples by solving the optimization problem defined in  \eqref{WHITE}. In the experiment, we
noticed that, under the white-box assumption, all the images can be perturbed into a target class and will not be detected by our detector when an unbounded perturbation size is allowed. 
There is no regularization term in  \eqref{WHITE}, but we use the Projected Gradient Descent (PGD) method \cite{Madry17}  to solve the optimization problem so that we can evaluate our method under a fixed perturbation size. Following the experiment in \cite{miller2019not}, performance measures were evaluated including the attack success rate and system defeat rate. An attack success means an image is perturbed such that it will be misclassified by the DNN classifier into the target class. System defeat 
is achieved when a perturbed image is misclassified {\em and} not detected by the detection method (Detection threshold is chosen with the false alarm rate 14\% like in \cite{miller2019not}). In Figure \ref{fig:WHITE_BOX}(a), we plot these measures under white-box (WB) and grey-box (GB) settings (grey-box is wherein the attacker knows the classifier but not the detector) versus 
perturbation size. The grey-box system defeat rate is always close to zero, which verifies the effectiveness of our detection method under the grey-box assumption. Since the AC-GAN discriminator has a DNN architecture, it is straightforward to compute the gradients and learn the perturbation by gradient descent; thus, an attack image that defeats the whole system can always be achieved when an unbounded perturbation size is allowed. However, to defeat the classifier and detector, a larger perturbation size is needed than to just defeat the DNN classifier. Also our detector is computationally much cheaper than the attacks -- an attack is made by hundreds of optimization steps, while our detection method (D-AD) only needs one forward pass to make a detection decision.

We also noticed that, with the same perturbation size, more optimization iterations are required to defeat the system (the classifier and the detector) than only to defeat the DNN classifier.  The average number of iterations required to defeat a classifier
or to defeat the system are reported in in Figure \ref{fig:WHITE_BOX} (b).

\subsection{Complexity}

The experiments in this work were performed on a computer with NVIDIA GeForce RTX 3090 GPU.  The GANs model training time, generator based detection time (GDT) and discriminator based detection time (DDT) are presented in Table
\ref{table:time}. While the GANs training time is high for Tiny-ImageNet-200, this time is comparable to that required for training the DNN.

\begin{table}[!ht]
\begin{center}
\scriptsize
\begin{tabular}{cccc}
\toprule
        
         & MNIST & CIFAR-10 & Tiny-ImageNet  \\ \hline \hline 

GAN training time& 4min 35s &1h 25min 57s& 15h 13min 2s  \\
GDT (per image)&  2.96s    & 12.9s   & n/a \\
DDT (per image) &  $2\times 10^{-5}$s    &    0.002s        & 0.03s          \\ 
\bottomrule
\end{tabular}
\caption{Training and detecting time complexity of our method on different datasets.}
\label{table:time}
\end{center}
\end{table}

\section{Conclusions}

In this paper, we proposed an AC-GAN based Anomaly detection method against TTE attacks (adversarial classifier inputs). We empirically showed that our method achieves state-of-the-art performance on two benchmark datasets, MNIST and CIFAR-10, considering both low and high confidence attacks. Our method can also be used to correct classifier decisions on detected attack examples.
Our results
indicate that, generally, internal-layer activations closer to the input
are  useful for detecting both high and low confidence
TTEs, compared with those closer to the output layer.  We also discussed the limitations of generator based detection and class correction methods.  
In future work, we may seek to improve the performance of generator based detection. 

\section{Acknowledgements}
This research was supported in part by AFOSR, ONR and NRC grants and
by a Cisco gift.

\bibliographystyle{plain}
\bibliography{gan-refs,adversarial,acgan-ada}

\begin{thebibliography}{10}

\bibitem{abusnaina2021adversarial}
Ahmed Abusnaina, Yuhang Wu, Sunpreet Arora, Yizhen Wang, Fei Wang, Hao Yang,
  and David Mohaisen.
\newblock Adversarial example detection using latent neighborhood graph.
\newblock In {\em Proceedings of the IEEE/CVF International Conference on
  Computer Vision}, pages 7687--7696, 2021.

\bibitem{akcay2018ganomaly}
Samet Akcay, Amir Atapour-Abarghouei, and Toby~P Breckon.
\newblock {GANomaly: Semi-supervised anomaly detection via adversarial
  training}.
\newblock In {\em Asian Conference on Computer Vision}, pages 622--637.
  Springer, 2018.

\bibitem{athalye2018obfuscated}
Anish Athalye, Nicholas Carlini, and David Wagner.
\newblock Obfuscated gradients give a false sense of security: Circumventing
  defenses to adversarial examples.
\newblock In {\em International conference on machine learning}, pages
  274--283. PMLR, 2018.

\bibitem{bai2020targeted}
Jiawang Bai, Bin Chen, Yiming Li, Dongxian Wu, Weiwei Guo, Shu-tao Xia, and
  En-hui Yang.
\newblock Targeted attack for deep hashing based retrieval.
\newblock In {\em ECCV}, 2020.

\bibitem{Openmax}
A.~Bendale and T.E. Boult.
\newblock Towards open set deep networks.
\newblock In {\em {Proc. CVPR}}, 2015.

\bibitem{cw}
N.~Carlini and D.~Wagner.
\newblock {Towards Evaluating the Robustness of Neural Networks}.
\newblock In {\em {Proc. IEEE Symposium on Security and Privacy}}, 2017.

\bibitem{carlini2017adversarial}
Nicholas Carlini and David Wagner.
\newblock Adversarial examples are not easily detected: Bypassing ten detection
  methods.
\newblock In {\em Proceedings of the 10th ACM workshop on artificial
  intelligence and security}, pages 3--14, 2017.

\bibitem{carlini2017towards}
Nicholas Carlini and David Wagner.
\newblock Towards evaluating the robustness of neural networks.
\newblock In {\em IEEE Symposium on Security and Privacy}, pages 39--57. IEEE,
  2017.

\bibitem{cohen2020detecting}
Gilad Cohen, Guillermo Sapiro, and Raja Giryes.
\newblock Detecting adversarial samples using influence functions and nearest
  neighbors.
\newblock In {\em Proceedings of the IEEE/CVF conference on computer vision and
  pattern recognition}, pages 14453--14462, 2020.

\bibitem{Smoothing19}
J.~Cohen, E.~Rosenfeld, and Z.~Kolter.
\newblock {Certified Adversarial Robustness via Randomized Smoothing}.
\newblock In {\em Proc. ICML}, 2019.

\bibitem{demontis2019adversarial}
Ambra Demontis, Marco Melis, Maura Pintor, Matthew Jagielski, Battista Biggio,
  Alina Oprea, Cristina Nita-Rotaru, and Fabio Roli.
\newblock {Why do adversarial attacks transfer? Explaining transferability of
  evasion and poisoning attacks}.
\newblock In {\em 28th USENIX Security Symposium}, pages 321--338, 2019.

\bibitem{imagenet}
J.~Deng, W.~Dong, R.~Socher, L.-J. Li, K.~Li, and L.~Fei-Fei.
\newblock {ImageNet: A Large-Scale Hierarchical Image Database}.
\newblock In {\em CVPR}, 2009.

\bibitem{dia2021localized}
Ousmane~Amadou Dia, Theofanis Karaletsos, Caner Hazirbas, Cristian~Canton
  Ferrer, Ilknur~Kaynar Kabul, and Erik Meijer.
\newblock Localized uncertainty attacks.
\newblock {\em arXiv preprint arXiv:2106.09222}, 2021.

\bibitem{AD}
R.~Feinman, R.~Curtin, S.~Shintre, and A.~Gardner.
\newblock {Detecting adversarial samples from artifacts}.
\newblock https://arxiv.org/abs/1703.00410v2, 2017.

\bibitem{feinman2017detecting}
Reuben Feinman, Ryan~R Curtin, Saurabh Shintre, and Andrew~B Gardner.
\newblock Detecting adversarial samples from artifacts.
\newblock {\em arXiv preprint arXiv:1703.00410}, 2017.

\bibitem{gans}
I.~Goodfellow, J.~Pouget-Abadie, M.~Mirza, B.~Xu, D.~Warde-Farley, S.~Ozair,
  A.~Courville, and Y.~Bengio.
\newblock Generative adversarial networks.
\newblock In {\em Proc. Neural Information Processing Systems (NIPS)}, 2014.

\bibitem{goodfellow2014explaining}
Ian~J Goodfellow, Jonathon Shlens, and Christian Szegedy.
\newblock Explaining and harnessing adversarial examples.
\newblock https://arxiv.org/abs/1412.6572, 2014.

\bibitem{Grosse}
K.~Grosse, P.~Manoharan, N.~Papernot, M.~Backes, and P.~Mc{D}aniel.
\newblock On the (statistical) detection of adversarial examples.
\newblock https://arxiv.org/pdf/1702.06280, 2017.

\bibitem{grosse2017statistical}
Kathrin Grosse, Praveen Manoharan, Nicolas Papernot, Michael Backes, and
  Patrick McDaniel.
\newblock On the (statistical) detection of adversarial examples.
\newblock {\em CoRR}, 2017.

\bibitem{hansen1990neural}
Lars~Kai Hansen and Peter Salamon.
\newblock Neural network ensembles.
\newblock {\em IEEE Transactions on Pattern Analysis and Machine Intelligence},
  12(10):993--1001, 1990.

\bibitem{he2016deep}
Kaiming He, Xiangyu Zhang, Shaoqing Ren, and Jian Sun.
\newblock Deep residual learning for image recognition.
\newblock In {\em Proceedings of the IEEE Conference on Computer Vision and
  Pattern Recognition}, pages 770--778, 2016.

\bibitem{hendrycks2016baseline}
Dan Hendrycks and Kevin Gimpel.
\newblock A baseline for detecting misclassified and out-of-distribution
  examples in neural networks.
\newblock In {\em {Proc. ICLR}}, 2017.

\bibitem{hu_new}
Shengyuan Hu, Tao Yu, Chuan Guo, Wei-Lun Chao, and Kilian~Q Weinberger.
\newblock A new defense against adversarial images: Turning a weakness into a
  strength.
\newblock {\em Advances in Neural Information Processing Systems}, 32, 2019.

\bibitem{ilyas2019adversarial}
Andrew Ilyas, Shibani Santurkar, Dimitris Tsipras, Logan Engstrom, Brandon
  Tran, and Aleksander Madry.
\newblock Adversarial examples are not bugs, they are features.
\newblock {\em arXiv preprint arXiv:1905.02175}, 2019.

\bibitem{isola2017image}
Phillip Isola, Jun-Yan Zhu, Tinghui Zhou, and Alexei~A Efros.
\newblock Image-to-image translation with conditional adversarial networks.
\newblock In {\em Proceedings of the IEEE conference on computer vision and
  pattern recognition}, pages 1125--1134, 2017.

\bibitem{kang2021rebooting}
Minguk Kang, Woohyeon Shim, Minsu Cho, and Jaesik Park.
\newblock Rebooting acgan: Auxiliary classifier gans with stable training.
\newblock {\em Advances in Neural Information Processing Systems}, 34, 2021.

\bibitem{karras2020analyzing}
Tero Karras, Samuli Laine, Miika Aittala, Janne Hellsten, Jaakko Lehtinen, and
  Timo Aila.
\newblock Analyzing and improving the image quality of stylegan.
\newblock In {\em Proceedings of the IEEE/CVF conference on computer vision and
  pattern recognition}, pages 8110--8119, 2020.

\bibitem{kingma2013auto}
D.P. Kingma and M.~Welling.
\newblock {Auto-Encodeing Variational Bayes}.
\newblock https://arxiv.org/abs/1312.6114.

\bibitem{krizhevsky2009learning}
Alex Krizhevsky.
\newblock Learning multiple layers of features from tiny images.
\newblock \url{http://www.cs.toronto.edu/~kriz/learning-features-2009-TR.pdf},
  2009.

\bibitem{lecun-mnisthandwrittendigit-2010}
Yann LeCun and Corinna Cortes.
\newblock {MNIST} handwritten digit database.
\newblock http://yann.lecun.com/exdb/mnist/, 2010.

\bibitem{lee2018simple}
Kimin Lee, Kibok Lee, Honglak Lee, and Jinwoo Shin.
\newblock A simple unified framework for detecting out-of-distribution samples
  and adversarial attacks.
\newblock {\em Advances in neural information processing systems}, 31, 2018.

\bibitem{Li_ICCV}
X.~Li and F.~Li.
\newblock Adversarial examples detection in deep networks with convolutional
  filter statistics.
\newblock In {\em {Proc. ICCV}}, 2017.

\bibitem{li2022semi}
Yiming Li, Baoyuan Wu, Yan Feng, Yanbo Fan, Yong Jiang, Zhifeng Li, and Shu-Tao
  Xia.
\newblock Semi-supervised robust training with generalized perturbed
  neighborhood.
\newblock {\em Pattern Recognition}, 124:108472, 2022.

\bibitem{pytorch-resnet}
Kuang Liu.
\newblock {Train CIFAR10 with PyTorch}.
\newblock https://github.com/kuangliu/pytorch-cifar.

\bibitem{Safety-net}
J.~Lu, T.~Issaranon, and D.~Forsyth.
\newblock Safetynet: detecting and rejecting adversarial examples robustly.
\newblock In {\em {Proc. ICCV}}, 2017.

\bibitem{ma2018characterizing}
Xingjun Ma, Bo~Li, Yisen Wang, Sarah~M Erfani, Sudanthi Wijewickrema, Grant
  Schoenebeck, Dawn Song, Michael~E Houle, and James Bailey.
\newblock Characterizing adversarial subspaces using local intrinsic
  dimensionality.
\newblock {\em arXiv preprint arXiv:1801.02613}, 2018.

\bibitem{Madry17}
A.~Madry, A.~Makelov, L.~Schmidt, D.~Tsipras, and A.~Vladu.
\newblock {Towards Deep Learning Models Resistant to Adversarial Attacks}.
\newblock In {\em Proc. ICLR}, June 2018.

\bibitem{madry2017towards}
Aleksander Madry, Aleksandar Makelov, Ludwig Schmidt, Dimitris Tsipras, and
  Adrian Vladu.
\newblock Towards deep learning models resistant to adversarial attacks.
\newblock {\em arXiv preprint arXiv:1706.06083}, 2017.

\bibitem{Metzen}
J.~Metzen, T.~Genewein, V.~Fischer, and B.~Bischoff.
\newblock On detecting adversarial perturbations.
\newblock In {\em {Proc. ICLR}}, 2017.

\bibitem{miller2019not}
David Miller, Yujia Wang, and George Kesidis.
\newblock {When not to classify: Anomaly detection of attacks (ADA) on DNN
  classifiers at test time}.
\newblock {\em Neural Computation}, 31(8):1624--1670, 2019.

\bibitem{miller2020adversarial}
David~J Miller, Zhen Xiang, and George Kesidis.
\newblock Adversarial learning targeting deep neural network classification: A
  comprehensive review of defenses against attacks.
\newblock {\em Proceedings of the IEEE}, 108(3):402--433, 2020.

\bibitem{mirza2014conditional}
Mehdi Mirza and Simon Osindero.
\newblock Conditional generative adversarial nets.
\newblock {\em arXiv preprint arXiv:1411.1784}, 2014.

\bibitem{miyato2018cgans}
Takeru Miyato and Masanori Koyama.
\newblock {cGANs with projection discriminator}.
\newblock {\em arXiv preprint arXiv:1802.05637}, 2018.

\bibitem{muller2019does}
Rafael M{\"u}ller, Simon Kornblith, and Geoffrey~E Hinton.
\newblock When does label smoothing help?
\newblock {\em Advances in neural information processing systems}, 32, 2019.

\bibitem{nesti2021detecting}
Federico Nesti, Alessandro Biondi, and Giorgio Buttazzo.
\newblock Detecting adversarial examples by input transformations, defense
  perturbations, and voting.
\newblock {\em IEEE Transactions on Neural Networks and Learning Systems},
  2021.

\bibitem{nguyen2015deep}
Anh Nguyen, Jason Yosinski, and Jeff Clune.
\newblock {Deep neural networks are easily fooled: High confidence predictions
  for unrecognizable images}.
\newblock In {\em Proceedings of the IEEE Conference on Computer Vision and
  Pattern Recognition}, pages 427--436, 2015.

\bibitem{odena2017conditional}
Augustus Odena, Christopher Olah, and Jonathon Shlens.
\newblock {Conditional image synthesis with auxiliary classifier GANS}.
\newblock In {\em International Conference on Machine Learning}, pages
  2642--2651. PMLR, 2017.

\bibitem{radford2015unsupervised}
Alec Radford, Luke Metz, and Soumith Chintala.
\newblock Unsupervised representation learning with deep convolutional
  generative adversarial networks.
\newblock {\em arXiv preprint arXiv:1511.06434}, 2015.

\bibitem{raghuram2020detecting}
Jayaram Raghuram, Varun Chandrasekaran, Somesh Jha, and Suman Banerjee.
\newblock {Detecting Anomalous Inputs to DNN Classifiers By Joint Statistical
  Testing at the Layers}.
\newblock {\em arXiv preprint arXiv:2007.15147}, 2020.

\bibitem{roth_odds}
Kevin Roth, Yannic Kilcher, and Thomas Hofmann.
\newblock The odds are odd: A statistical test for detecting adversarial
  examples.
\newblock In {\em International Conference on Machine Learning}, pages
  5498--5507. PMLR, 2019.

\bibitem{samangouei2018defense}
Pouya Samangouei, Maya Kabkab, and Rama Chellappa.
\newblock {Defense-GAN: Protecting classifiers against adversarial attacks
  using generative models}.
\newblock {\em arXiv preprint arXiv:1805.06605}, 2018.

\bibitem{schlegl2019f}
Thomas Schlegl, Philipp Seeb{\"o}ck, Sebastian~M Waldstein, Georg Langs, and
  Ursula Schmidt-Erfurth.
\newblock {f-AnoGAN: Fast unsupervised anomaly detection with generative
  adversarial networks}.
\newblock {\em Medical Image Analysis}, 54:30--44, 2019.

\bibitem{schlegl2017unsupervised}
Thomas Schlegl, Philipp Seeb{\"o}ck, Sebastian~M Waldstein, Ursula
  Schmidt-Erfurth, and Georg Langs.
\newblock Unsupervised anomaly detection with generative adversarial networks
  to guide marker discovery.
\newblock In {\em International Conference on Information Processing in Medical
  Imaging}, pages 146--157. Springer, 2017.

\bibitem{Schwarz}
G.~Schwarz.
\newblock Estimating the dimension of a model.
\newblock {\em The Annals of Statistics}, 6(2):461--464, 1978.

\bibitem{sun2020towards}
Jiachen Sun, Yulong Cao, Qi~Alfred Chen, and Z~Morley Mao.
\newblock Towards robust $\{$LiDAR-based$\}$ perception in autonomous driving:
  General black-box adversarial sensor attack and countermeasures.
\newblock In {\em 29th USENIX Security Symposium (USENIX Security 20)}, pages
  877--894, 2020.

\bibitem{tian2021detecting}
Jinyu Tian, Jiantao Zhou, Yuanman Li, and Jia Duan.
\newblock Detecting adversarial examples from sensitivity inconsistency of
  spatial-transform domain.
\newblock In {\em Proceedings of the AAAI Conference on Artificial
  Intelligence}, 2021.

\bibitem{tramer2021detecting}
Florian Tramer.
\newblock Detecting adversarial examples is (nearly) as hard as classifying
  them.
\newblock In {\em ICML 2021 Workshop on Adversarial Machine Learning}, 2021.

\bibitem{tramer2017ensemble}
Florian Tram{\`e}r, Alexey Kurakin, Nicolas Papernot, Ian Goodfellow, Dan
  Boneh, and Patrick McDaniel.
\newblock {Ensemble adversarial training: Attacks and defenses}.
\newblock {\em arXiv preprint arXiv:1705.07204}, 2017.

\bibitem{tsipras2018robustness}
Dimitris Tsipras, Shibani Santurkar, Logan Engstrom, Alexander Turner, and
  Aleksander Madry.
\newblock Robustness may be at odds with accuracy.
\newblock {\em arXiv preprint arXiv:1805.12152}, 2018.

\bibitem{AC-GAN-ada-github}
Hang Wang, Zhen Xiang, David~J. Miller, and George Kesidis.
\newblock {AC-GAN-ADA for Detecting Adversarial Examples}.
\newblock \url{https://github.com/wanghangpsu/acgan-ada}, 2022.

\bibitem{xiang2020detection}
Zhen Xiang, David~J Miller, and George Kesidis.
\newblock Detection of backdoors in trained classifiers without access to the
  training set.
\newblock {\em IEEE Transactions on Neural Networks and Learning Systems},
  2020.

\bibitem{xiang2022post}
Zhen Xiang, David~J Miller, and George Kesidis.
\newblock Post-training detection of backdoor attacks for two-class and
  multi-attack scenarios.
\newblock {\em arXiv preprint arXiv:2201.08474}, 2022.

\bibitem{yin2019gat}
Xuwang Yin, Soheil Kolouri, and Gustavo~K Rohde.
\newblock Gat: Generative adversarial training for adversarial example
  detection and robust classification.
\newblock In {\em International conference on learning representations}, 2019.

\bibitem{zenati2018efficient}
Houssam Zenati, Chuan~Sheng Foo, Bruno Lecouat, Gaurav Manek, and
  Vijay~Ramaseshan Chandrasekhar.
\newblock {Efficient GAN-based anomaly detection}.
\newblock {\em arXiv preprint arXiv:1802.06222}, 2018.

\bibitem{zheng2018robust}
Zhihao Zheng and Pengyu Hong.
\newblock Robust detection of adversarial attacks by modeling the intrinsic
  properties of deep neural networks.
\newblock In {\em Proceedings of the 32nd International Conference on Neural
  Information Processing Systems}, pages 7924--7933, 2018.

\end{thebibliography}

\appendix
\section{Example attack images of different attack method}
\label{appendix:attack-images}
\begin{figure}[!ht]

  \centering
  \includegraphics[width=12cm]{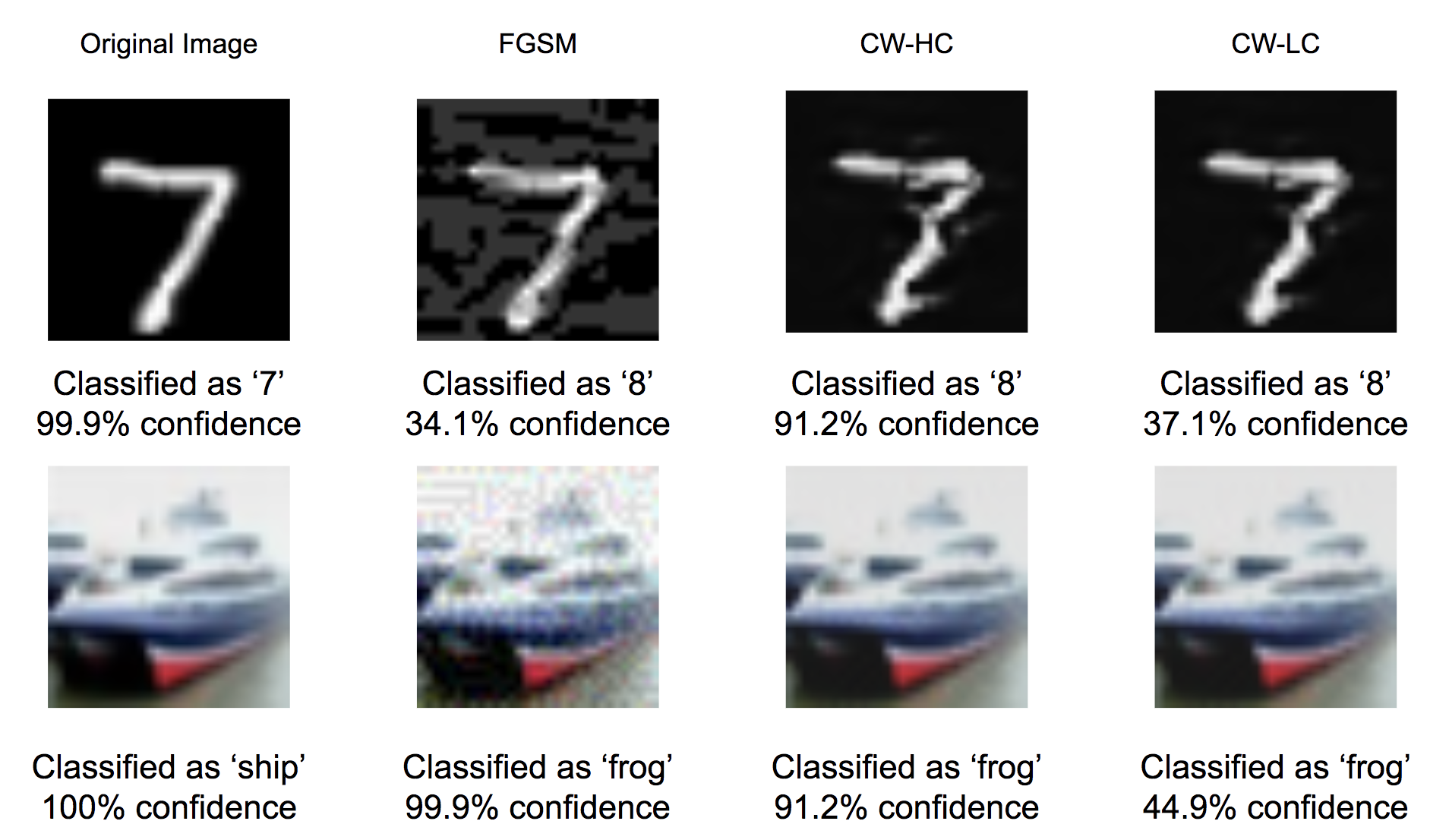}
  \caption{Examples of clean images, FGSM attack images, CW high confidence (CW-HC) images and CW low confidence (CW-LC) images from MNIST (first row) and CIFAR-10 (second row) dataset. }
\label{fig:attack-examples}

\end{figure}

Figure \ref{fig:attack-examples} shows some examples of attack images from MNIST and CIFAR-10 dataset.

\section{AC-GAN model architectures}
\label{appendix:architecture}

\begin{figure}[!ht]
\begin{minipage}[b]{.48\linewidth}
  \centering
  \centerline{\includegraphics[width=5cm]{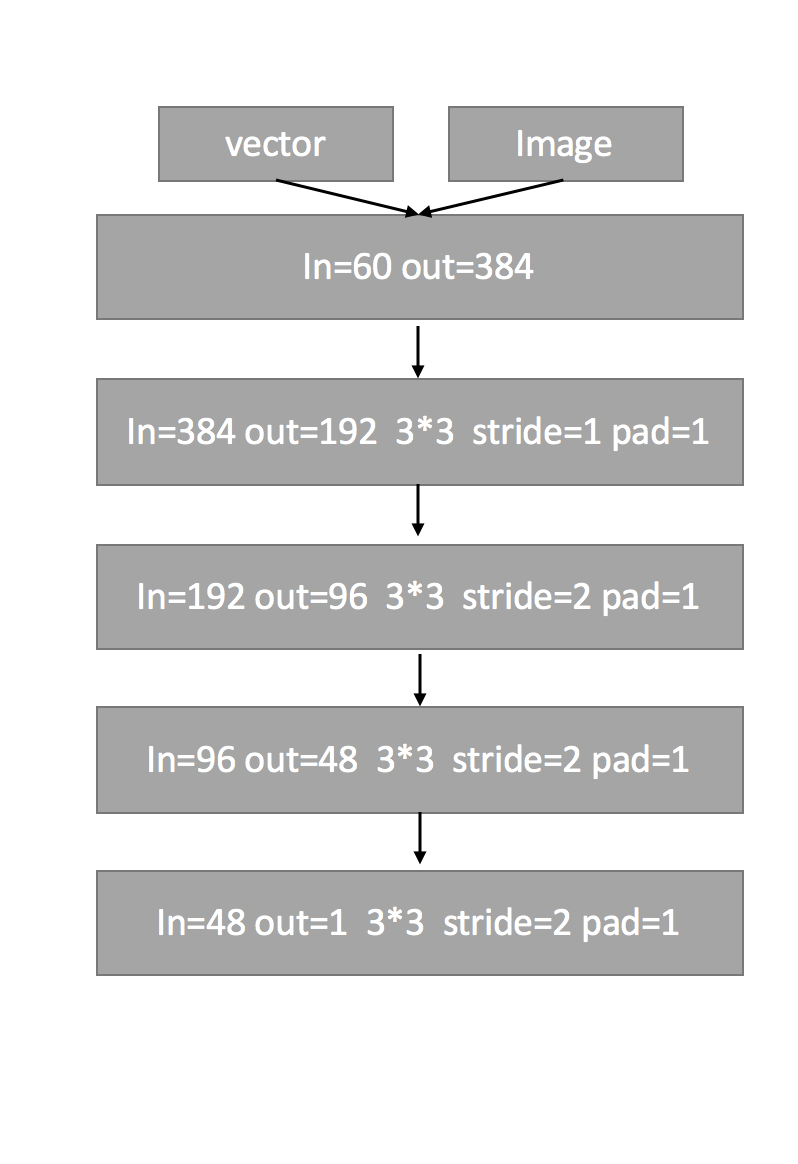}}
  \centerline{(a) Generator}
\end{minipage}
\begin{minipage}[b]{0.48\linewidth}
  \centering
  \centerline{\includegraphics[width=5cm]{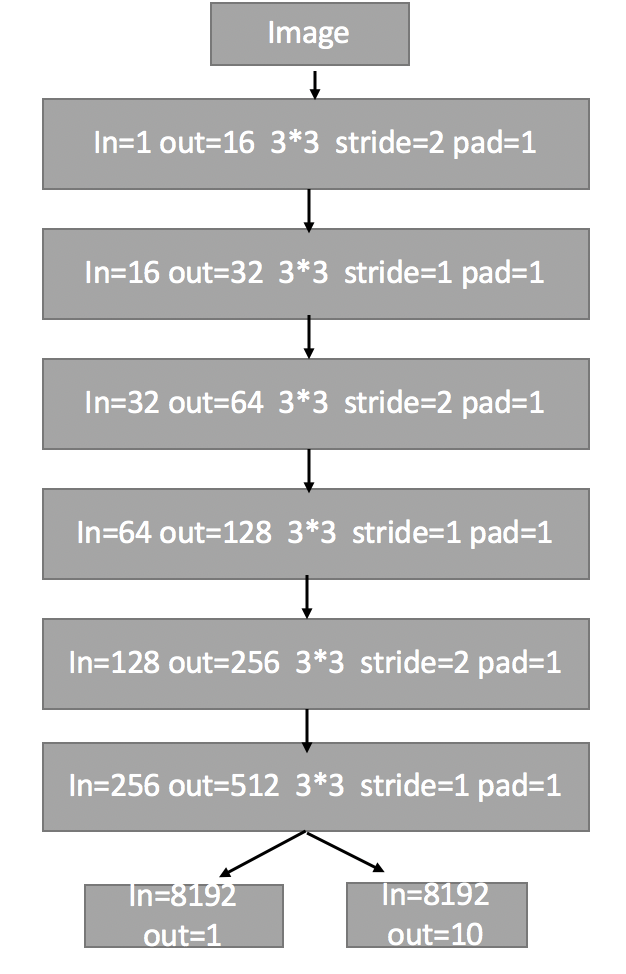}}
  \centerline{(b) Discriminator}
\end{minipage}
\caption{The structure of the Generator and the Discriminator of AC-GAN model used for MNIST dataset. For the generator, the block with only `in' and `out' represents the fully connected layer and the block with `stride' and `padding' represents the transposed convolutional layer. For the discriminator the block with only `in' and `out' represents the fully connected layer and the block with `stride' and `padding' represents the convolutional layer }
\label{fig:strucmnist}
\end{figure}

For the MNIST dataset, a simple AC-GAN architecture can capture the training distribution. The generator and the discriminator architectures are shown in Figure \ref{fig:strucmnist}. Note that in the discriminator, there are linear layers and convolutional layers. Each convolutional layer is followed by a batch normalization layer, a ReLU layer and a drop-out layer with dropping rate 0.5. 
In the generator, there are linear layers and deconvolutional layers. Each deconvolutional layer is followed by a batch normalization layer and a ReLU layer. 

\begin{figure}[!ht]
  \centering
  \includegraphics[width=10cm]{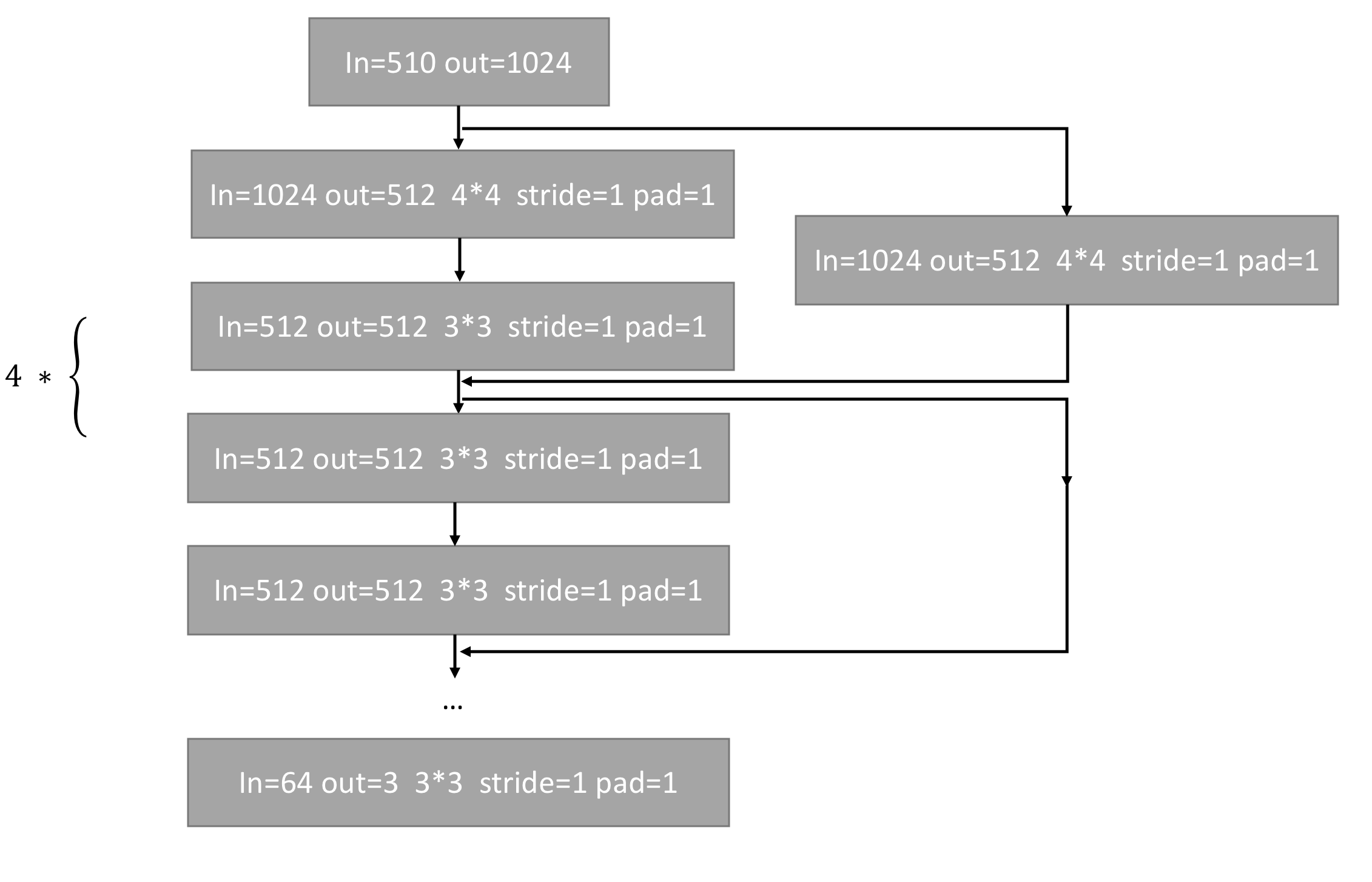}
  \caption{Generator used in AC-GAN for CIFAR-10 dataset. The block with only `in' and `out' represents the fully connected layer and the block with `stride' and `padding' represents the transposed convolutional layer}
\label{fig:resg}
\end{figure}

For the CIFAR-10 and Tiny-ImageNet-200 dataset, the architectures described in Figure \ref{fig:strucmnist} are not complex enough to capture the data distribution, So for the discriminator, we borrow the ResNet-18 architecture from \cite{he2016deep}. In the GANs model the generator's complexity and the discriminator's complexity should match; based on the ResNet-18 architecture, we used a transposed convolutional based ResNet-like generator. The architecture of the generator is shown in Figure \ref{fig:resg}.

\section{ADA revisited: Auto-encoder based Dimension Reduction}
\label{appendix:a}
For the ADA method, it is challenging to model the distribution of high-dimensional features in the internal layers, mainly because BIC based model order selection \cite{Schwarz} may grossly underestimate the number of Gaussian components in a Gaussian mixture for modeling these high-dimensional features. In this section, we introduce an auto-encoder based dimension reduction method to mitigate this problem.  This method was used to obtain the ADA detection results reported in the main paper. Consider the DNN network as a cascade
of two sub-networks $h_{\rm pre}$ and $h_{\rm post}$. Give an input image $x$, $z = h_{\rm pre}(x)$ is an internal layer feature vector and $h_{\rm post}(h_{\rm pre}(x))=h_{\rm post}(z)$ is the network's output. Then an auto-encoder model can be used to reconstruct the internal layer features as $g_{\rm dec}(g_{\rm enc}(z))$, where $g_{\rm enc}$ is the encoder network and $g_{\rm dec}$ is the decoder network. The training objective of the auto-encoder is as follows:
\begin{equation}
    L_{\rm auto} = L_{\rm ce}\{ h_{\rm post}(g_{\rm dec}(g_{\rm enc}(h_{\rm pre}(x)))), y\} +\lambda \|h_{\rm pre}(x) - g_{\rm dec}(g_{\rm enc}(h_{\rm pre}(x))) \|^2,\label{ada_net}
\end{equation}
where \gk{$L_{\rm ce}$ is the cross entropy loss,
$x\sim p_{\rm data}$, and $y$ is $x$'s class label.}
\hw{We noticed that ADA's performance is not sensitive to the choice of $\lambda$. 
Thus, we simply used $\lambda=1$ for all datasets}.
The parameters of the DNN classifier are fixed during the training process of the auto-encoder. After training, the encoder $g_{\rm enc}$ gives a good low dimension representation of the high-dimensional features in the DNN's internal layer.  These low-dimensional features are used for ADA based anomaly detection.

\section{A pix2pix based robust image classification model}
\label{appendix:b}

In this section, we discuss an improved Defense-GAN using image to image translation. Note that  when given an image $x$ to be classified, Defense-GAN first searches in the input space to find a vector $z^*$ that minimizes the distance between the generated image $G(z)$ and $x$; then the optimized vector $z^*$ is fed into the Generator to create a reconstructed image $x^* = G(z^*)$ which, instead of $x$, is fed into the DNN classifier to achieve robust classification. Since the GAN is trained on the clean image dataset, the Generator tends to produce images following the clean distribution.
Thus, the reconstructed image may eliminate the perturbations of an attack image while preserving the features essential to correct classification.

However, 1) the reconstruction is an optimization process and requires many iterations to produce an image with low reconstruction error, and
2) the reconstruction is hard to perform since the image should be generated from a random vector $z$, see Figure \ref{fig:images}. In our work, \hw{we modified the pix2pix method \cite{isola2017image} to essentially act as an auto-encoder, see below. This approach makes the reconstruction task much easier by inputting the image $x$ into the Generator to produce a reconstructed image $x_r = G(x)$. The discriminator, when given an image $x_r$, outputs probability $D(x_r, x)$ that $x_r$ is real given $x$}. Although the Variant Auto-Encoder (VAE)\cite{kingma2013auto} model is also used for image to image translation, without the adversarial training process and without a Discriminator model to penalize deviations from a 
clean 
image, it can only reconstruct a given image and cannot eliminate adversarial perturbations. 

\begin{figure}[!ht]
\begin{minipage}[b]{1\linewidth}
  \centering
  \centerline{\includegraphics[width=10cm]{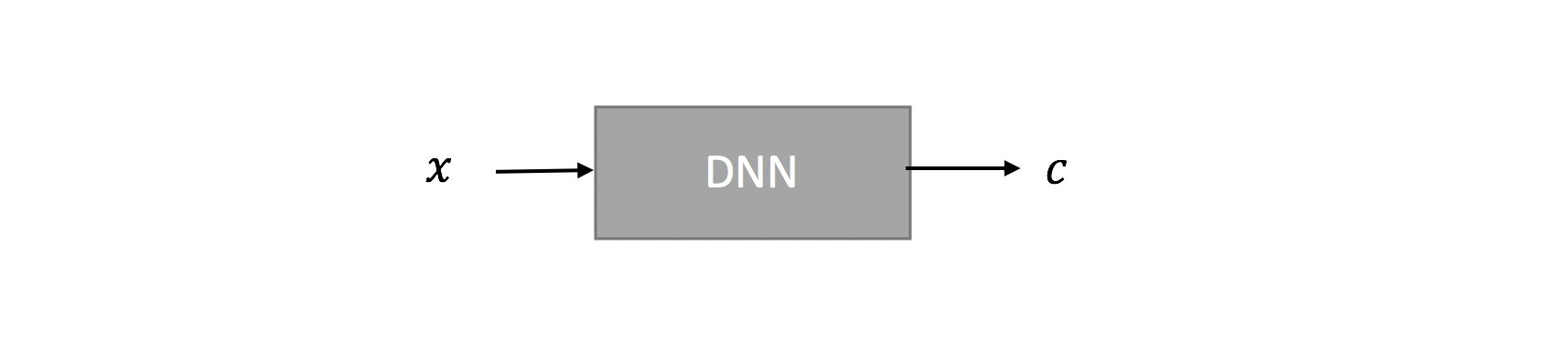}}
  \centerline{(a) non-robust classification}
\end{minipage}
\begin{minipage}[b]{1\linewidth}
  \centering
  \centerline{\includegraphics[width=10cm]{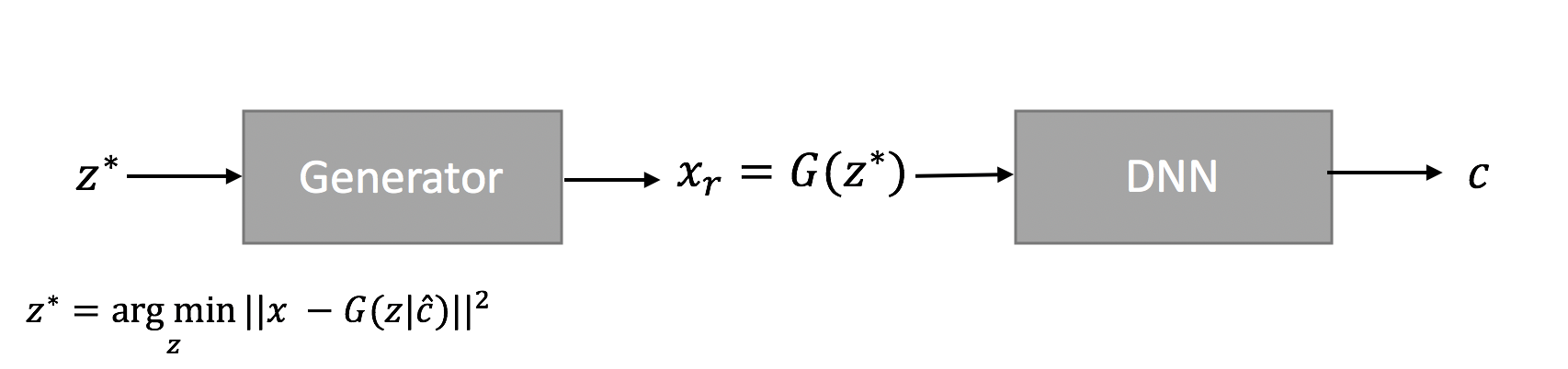}}
  \centerline{(b) Defense-GAN}
\end{minipage}
\begin{minipage}[b]{1\linewidth}
  \centering
  \centerline{\includegraphics[width=10cm]{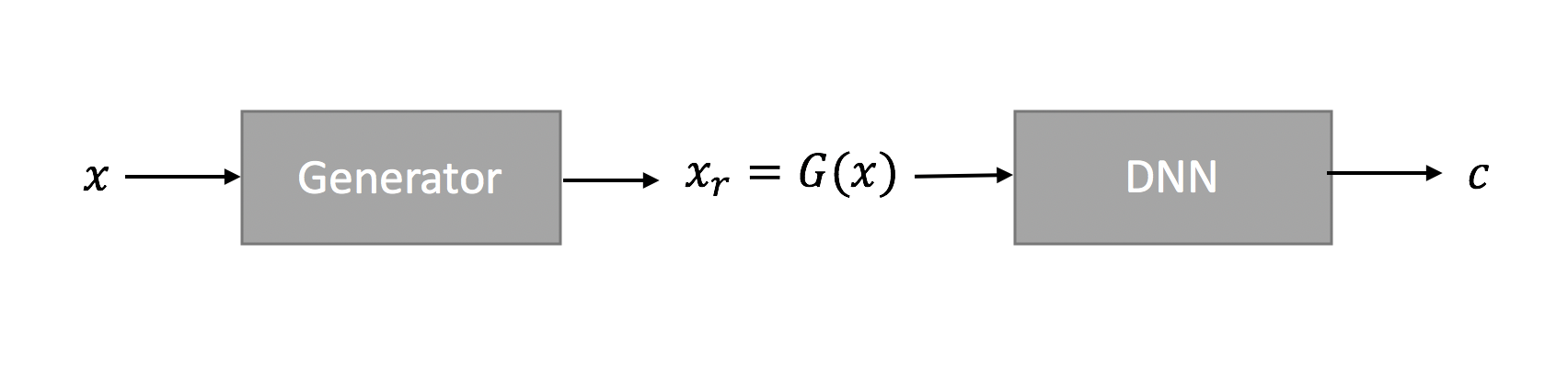}}
  \centerline{(c) \gk{pix2pix}}
\end{minipage}
\caption{The difference between the Defense-GAN method and pix2pix.}
\label{fig:compare}
\end{figure}

\begin{figure}[!ht]

  \centering
  \includegraphics[width=12cm]{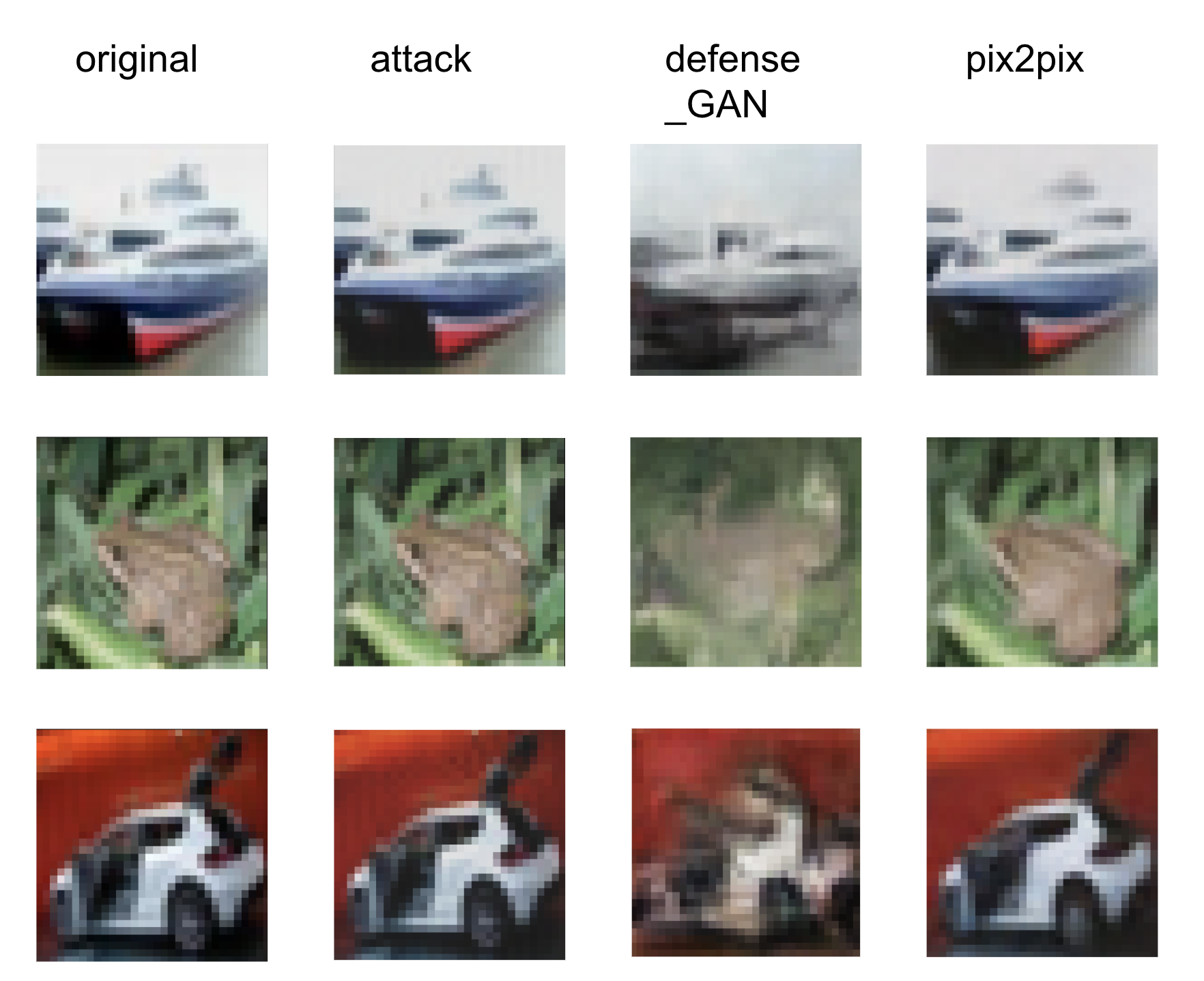}
  \caption{Comparison between the Defense-GAN reconstruction and our pix2pix reconstruction images. Images in the first column are original images; images in the second column are the corresponding attack images created by the CW-targeted high-confidence attack. In the third column, there are images reconstructed from the attack images using Defense-GAN. Images in the fourth column are the reconstructions of attack images using pix2pix. The perturbations in attack images are imperceptible. The quality of the defense-GAN reconstruction is low and features necessary for classification are not well-preserved. Pix2pix's  reconstruction quality is much higher.}
\label{fig:images}

\end{figure}

\subsection{Model Learning}
\label{sec:model_learn}

To learn a pix2pix model, we need a dataset with image pairs $\{x_a, x_c\}$. For the class label correction scenario, we want to use the Generator to reconstruct a clean image from an attack image. In an image pair, $x_c$ should be a clean image, and $x_a$ should be the corresponding attack image. It would be easy to generate $x_a$ if we know the attack method. But we are considering the case when the attack method is not known; also we want our model to be robust to all kinds of TTE attacks. For $x_a$, we use the noisy version of a clean image $x_c$ to imitate an attack image:
\begin{equation}
    x_a = x_c + \nu,
\end{equation}
where $x_c$ is an image directly from the training set and $\nu$ is a 
\gk{uniformly} random noise image. \gk{In our experiments, for both MNIST and CIFAR-10, performance was not sensitive to the variance of the random noise $\nu$, and we herein report results only for variance 0.05. A relatively small variance or even 0 variance (i.e., when $x_a = x_c$) gives comparable performance.}
Our pix2pix model is trained based on the following objective function:
\begin{equation}
  \min_G \max_D  \E \log D(x_c+\nu,x_c)
+  \E \log(1 - D(x_c+\nu, G(x_c+\nu))
+ \lambda \E\|x_c - G(x_c+\mu)\|^2. 
\end{equation}
where $x_c\sim p_{\rm data}$ and $\nu\sim$uniform are independent,
and parameter $\lambda>0$. \hw{Note that the discriminator now takes two images as input, the reference image $x_c+\nu$ and the image to be discriminated (which can be a clean training $x_c$ or a generated image $G(x_c+\mu)$). The output of the discriminator is the probability that the image ($x_c$ or $G(x_c+\mu)$) is real given the reference image.}



\subsection{Test Inference}
When given an image $x$, a reconstructed image will be generated from the trained Generator $G^*$, i.e., $x_r = G^*(x)$, and used for classification (see Figure \ref{fig:compare}). Since the Generator is trained to generate images following the clean data distribution \gk{as well as close to the input image}, the reconstruction \gk{is expected to} preserve the clean features that are essential for classification while filtering out perturbations that were added to form an attack image.

\subsection{Reconstruction Error}
We first compare the image reconstruction error between our
\gk{AC-GAN based} method and Defense-GAN. Figure \ref{fig:images} shows some examples of the defense-GAN reconstruction images and the pix2pix reconstruction images. For the defense-GAN reconstructed images, the reconstruction error is high and necessary features are not preserved -- we cannot recognize the object when shown the images.  However, for the pix2pix model, the reconstructed images are easy to recognize. We report the mean-squared reconstruction error for the Defense-GAN method and our method on different datasets under different attacks, shown in Table \ref{table:recon_error}.

 \begin{table}[!ht]
\begin{center}
\scriptsize
\begin{tabular}{ccccc}
\toprule
         & \multicolumn{2}{c}{MINST}         & \multicolumn{2}{c}{CIFAR-10}      \\ \cmidrule(lr){2-3}\cmidrule(lr){4-5} 
         & CW-HC & CW-LC & CW-HC & CW-LC \\ \hline \hline 

D-GAN\cite{samangouei2018defense} &   0.0136   &   0.0131        &0.0102        &0.0098            \\

pix2pix  &        {\bf 0.0098}  &   {\bf 0.0086}          &    {\bf 0.00106}    &  {\bf 0.00095}         \\             
\bottomrule
\end{tabular}
\caption{Comparison of mean-squared reconstruction errors for defense-GAN and our pix2pix method. CW-HC means CW high confidence attack; CW-LC means CW low confidence attack. D-GAN means the Defense-GAN method.}
\label{table:recon_error}
\end{center}
\end{table}

We also evaluate the performance of class-label correction for our method, compared with Defense-GAN. The results, which demonstrate the superiority of our method, are shown in Table \ref{table:robust}.
\end{document}